\documentclass{article}

\usepackage{floatrow}
\newfloatcommand{capbtabbox}{table}[][\FBwidth]

\usepackage{pifont}
\usepackage{booktabs}
\usepackage{amsmath}
\usepackage{amssymb}
\usepackage{balance}
\usepackage{multicol}
\usepackage{multirow}
\usepackage{xcolor}
\usepackage{xspace}
\usepackage{colortbl}
\usepackage{wrapfig} 
\definecolor{mygray}{gray}{.9}
\usepackage{caption} 
\usepackage[font=footnotesize,skip=3pt,subrefformat=parens]{subcaption}
\usepackage{graphicx}
\usepackage{makecell}
\usepackage{enumitem} 

\usepackage{bm}
\usepackage{xcolor}
\definecolor{iccvblue}{rgb}{0.21,0.49,0.74}
\usepackage[pagebackref,breaklinks,colorlinks,allcolors=iccvblue]{hyperref}
\usepackage[capitalize]{cleveref} 

    \usepackage[final]{neurips_2025}

\title{EgoDTM: Towards 3D-Aware Egocentric Video-Language Pretraining}
\author{Boshen Xu$^1$\quad Yuting Mei$^{1}$\quad Xinbi Liu$^{1}$\quad Sipeng Zheng$^{2}$\quad Qin Jin$^{1}$\thanks{Qin Jin is the corresponding author.}\\
$^1$ AIM3 Lab, Renmin University of China\quad
$^2$ BeingBeyond\\
}

\begin{document}

\maketitle
\begin{abstract}
Egocentric video-language pretraining has significantly advanced video representation learning.
Humans perceive and interact with a fully 3D world, developing spatial awareness that extends beyond text-based understanding.
However, most previous works learn from 1D text or 2D visual cues, such as bounding boxes, which inherently lack 3D understanding. 
To bridge this gap, we introduce \textbf{EgoDTM}, an \textbf{Ego}centric \textbf{D}epth- and \textbf{T}ext-aware \textbf{M}odel, jointly trained through large-scale 3D-aware video pretraining and video-text contrastive learning.
EgoDTM incorporates a lightweight 3D-aware decoder to efficiently learn 3D-awareness from pseudo depth maps generated by depth estimation models.
To further facilitate 3D-aware video pretraining, we enrich the original brief captions with hand-object visual cues by organically combining several foundation models. 
Extensive experiments demonstrate EgoDTM's superior performance across diverse downstream tasks, highlighting its superior 3D-aware visual understanding. 
Code: \url{https://github.com/xuboshen/EgoDTM}.

\end{abstract}

\section{Introduction}
\label{sec:intro}

\begin{wrapfigure}{r}{0.5\textwidth}
  \centering
    \includegraphics[width=\linewidth]{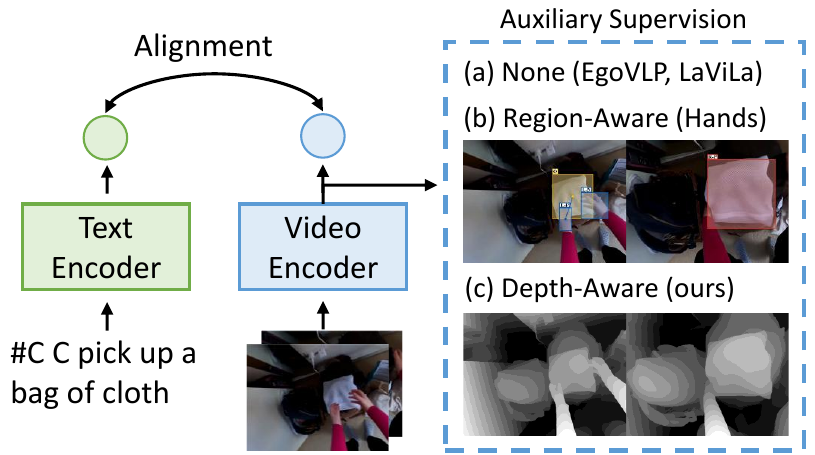}
  \caption{Comparison of egocentric pretraining paradigms. While previous paradigms focus on text-based~\cite{lin2022egovlp,pramanick2023egovlpv2,lavila} or 2D spatial region-aware learning~\cite{zhang2023helping}, EgoDTM incorporates 3D spatial information to enhance video representations.}
  \label{fig:motivation}
\end{wrapfigure}

The development of embodied AI capable of fulfilling diverse societal roles has long been a goal in artificial intelligence~\cite{hu2023robotsurvey,plizzari2023outlook}.
A promising path to achieving this vision involves developing egocentric AI, which can comprehend human activities by analyzing large-scale egocentric videos captured from a first-person perspective using wearable devices.
These videos provide rich insights into hand-object interactions (HOI)~\cite{fan2023arctic,zhan2024oakink2,epic-kitchens-100,liu2022hoi4d}, offering a window into how individuals interact with nearby objects through their hands.
With the emergence of large-scale egocentric datasets~\cite {grauman2022ego4d}, video-language pretraining~\cite{lin2022egovlp,lavila} has become a dominant paradigm for learning egocentric video representations, significantly improving performance on downstream tasks such as video-text retrieval~\cite{epic-kitchens-100,actor_observer} and action recognition~\cite{charades-ego,egtea}. 

Humans possess an innate ability to perceive and reason about 3D spatial relationships, effortlessly perceiving relative distances and spatial arrangements from visual cues alone~\cite{spelke2010beyond,koenderink1992surface,shepard1971mental}.
While there is no universal definition of a model's 3D awareness, we define it as the ability to infer 3D information from 2D images. 
However, models pretrained solely on 2D frames and general-purpose text (e.g., CLIP~\cite{clip}) often struggle to develop robust 3D awareness~\cite{cvpr-2024probing3D,nips-man2024lexicon3d,ramakrishnan2024doesspatialcognitionemerge}. 
To equip egocentric models with such 3D awareness, we wonder: Can video-language models better understand egocentric contexts by incorporating 3D-aware perception?

Achieving this goal presents two key challenges.
First, 3D representations are diverse, and obtaining corresponding labels is often costly.
Recent works~\cite{yue2024fit3d,zhang2024condense} explore using multi-view images for 3D reconstruction to distill 3D awareness into 2D models. However, in egocentric scenarios, 
depth maps are more practical and easier to obtain.
Depth maps provide direct 3D distance information and help distinguish salient objects from the background, offering crucial cues for spatial understanding.
Unfortunately, existing egocentric datasets with depth maps~\cite{liu2022hoi4d,ragusa2021meccano,Kwon2021h2o} remain limited in both scale and diversity for large-scale pretraining.  
Recent foundation models for depth estimation~\cite{depthanything,depthanythingv2,iclr2025depthpro} demonstrate strong out-of-domain generalization, enabling us to construct a large-scale depth-augmented egocentric dataset at a low cost. 

Second, effective 3D-aware pretraining requires bridging the modality gap between depth and text.
As illustrated in~\Cref{fig:motivation},
common video-language pretraining~\cite{lin2022egovlp,pramanick2023egovlpv2,lavila} primarily relies on textual supervision, thus avoiding this challenge. 
Recent works~\cite{zhang2023helping,phan2024henasy} explore region-aware video-language pretraining using non-pixel-level cues, such as text and sparse object bounding boxes,  which face only minor modality gaps. 
However, unlike these non-pixel-level cues, regressing the pixel-level outputs (e.g., depth maps) requires fundamentally different capabilities compared to predicting texts or bounding boxes, as suggested by previous studies~\cite{li2023maskdino,li2023flip}.
Additionally, depth estimation typically requires high-resolution inputs and multi-scale features, whereas video-language models rely on large batch sizes for contrastive learning, making direct integration nontrivial. Thus, designing an effective learning approach and enriching textual supervision with spatial information are crucial for successful 3D-aware pretraining.

To address these aforementioned challenges, we introduce \textbf{EgoDTM}, a novel 3D-aware egocentric video-language model that learns video representation from depth maps and spatially informed captions. 
In addition to dual transformer encoders for video-text alignment, EgoDTM adopts a 3D-aware module for video pretraining and a data construction pipeline to enrich captions with spatial information.
To adapt the 3D-aware module for the video-language framework, we propose a unified visual representation with a lightweight depth decoder, supervised by depth predictions from foundation models~\cite{depthanything}.
Specifically, the lightweight depth decoder takes as input the visual representation from the last layer of the video encoder and predicts the composition of discrete adaptive bins~\cite{bhat2021adabins,bhat2022localbins,shao2023iebins,li2024binsformer} to estimate low-resolution depth maps.
Moreover,  
we adopt off-the-shelf visual foundation models~\cite{hoidetector,sam2} to create high-quality HOI masks by first detecting HOIs, then selecting key frames, and finally tracking bidirectionally. To enrich the original captions with spatial information, we leverage a large language model (LLM) guided by the temporally consistent HOI masks.
Through 3D-aware video-language pretraining, EgoDTM improves visual generalization on downstream tasks involving egocentric HOIs.

Our contributions are threefold:
\noindent(1) We introduce EgoDTM, a 3D-aware egocentric video-language model learned from 3D-aware video-language pretraining.
\noindent(2) We develop a lightweight 3D-aware decoder for depth estimation and a data construction pipeline to enrich captions with spatial information. As a byproduct, we generate millions of egocentric data, including captions, HOI boxes, HOI masks, and depth maps.
\noindent(3) Extensive experiments demonstrate that EgoDTM significantly enhances performance on video understanding tasks like video-text matching, and 3D understanding tasks like robot manipulation.

\section{Related Works}
\label{sec:related_works}

\noindent\textbf{Egocentric Video-Language Pretraining.} 
Egocentric video understanding~\cite{grauman2022ego4d,openset-nips2023,plizzari2023can,zhang2022fine,mangalam2024egoschema,xu2023egopca,epic-kitchens-100,xu2023pov,xue2023learning,li2021egoexo,xu2025do} typically involves human daily interaction between hands and objects from a first-person perspective.
Inspired by visual-language pretraining paradigms in third-person datasets~\cite{bain2021frozen,luo2022clip4clip,clip,phan2024henasy,zhang2023helping}, EgoVLP~\citep{lin2022egovlp} firstly proposes to conduct egocentric video-language pretraining on the large-scale egocentric dataset Ego4D~\cite{grauman2022ego4d}, which aims to learn video representations from massive video-text data via contrastive learning.
Since EgoVLP, similar paradigms~\cite{lin2022egovlp,pramanick2023egovlpv2,lavila,ashutosh2023hiervl} have gained large success in EgoHOI understanding, 
For example, LaViLa~\cite{lavila} employs a video-conditioned GPT-2~\citep{gpt2} and a T5~\citep{t5} rephraser to generate text descriptions for egocentric videos to expand video-text data.
Despite the progress, pretraining with text alone often lacks precision in target localization. 
In response, some studies have focused on developing region-aware representations for EgoVLMs. 
For example, HelpingHands~\cite{zhang2023helping} proposes learning from noisy hand-object detection results, while HENASY~\cite{phan2024henasy} ensembles an additional hierarchical encoder to learn HOI region-aware representation. 
However, these methods are limited to 2D reasoning and lack an understanding of real-world 3D context.
In this work, we take a step towards 3D-aware egocentric video-language pretraining.

\begin{figure*}[t]
  \centering
\includegraphics[width=\linewidth]{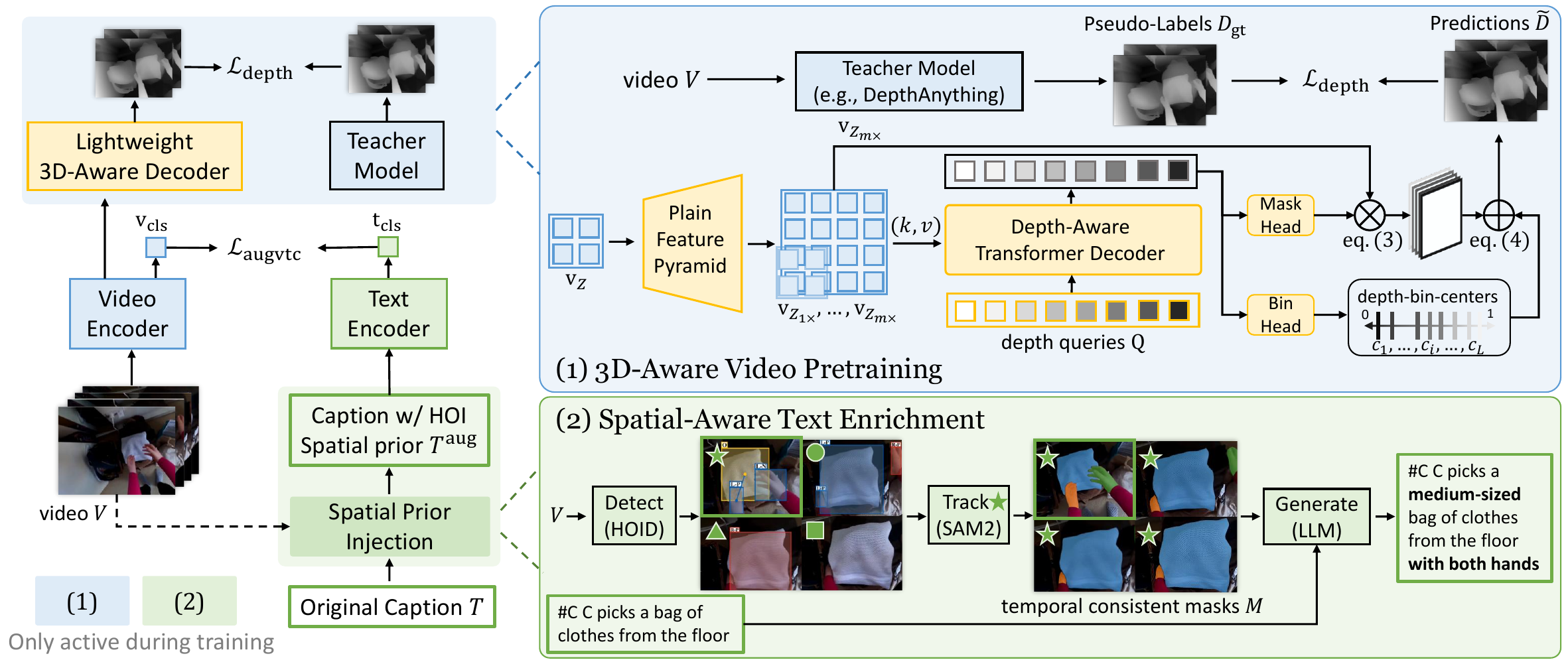}  
  \caption{
  EgoDTM learns 3D-aware representations from depth and text. Our dual encoders are constructed using only transformers~\cite{dosovitskiy2020image,vaswani2017attention,clip} with flash attention~\cite{dao2022flashattention}. During pretraining, we conduct (1) 3D-aware video pretraining: we design a lightweight 3D-aware decoder to predict depth using visual feature maps, supervised by a teacher foundation model~\cite{depthanythingv2}. The decoder contains a plain feature pyramid to get multi-scale features, a depth-aware transformer decoder to process depth queries with video features, and the heads to predict depth maps; (2) Spatial-aware textual enrichment: we enhance captions with spatial information by organically combining foundation models in the detect-track-generate pipeline. Different green markers denote inconsistency of HOI predictions; identical ones indicate consistency.
  }
  \label{fig:arch}
\end{figure*}

\noindent\textbf{3D-Awareness of 2D Vision Models.}
As suggested by research in developmental psychology and psychophysics, we humans are capable of understanding 3D information, like depth and orientations, from only 2D visual signals~\cite{spelke2010beyond,koenderink1992surface,shepard1971mental}.
With the growing interest in embodied intelligence~\cite{hu2023robotsurvey,mccarthy2024survey2} and personal AI assistants~\cite{plizzari2023outlook}, the demand for 3D-aware 2D visual models has become increasingly important. 
Although there is no widely accepted consensus to define and build effective 3D-aware models, some studies have started to assess the 3D awareness of visual foundation models.
For example, Probing3D~\cite{cvpr-2024probing3D} demonstrated that CLIP, which is pretrained exclusively with text, is significantly limited in predicting depth and 3D surface normals. 
EysWideShut\cite{tong2024eyes} revealed that CLIP struggles with 3D-related tasks, such as recognizing object orientation and understanding compositional contexts.
More recently, relevant works have made endeavors to enable visual foundation models (e.g., CLIP) with 3D awareness by fine-tuning on 3D-aware data.
For instance, FiT3D~\cite{yue2024fit3d} finetunes visual foundation models with 3D Gaussian features reconstructed from multi-view images.
SpatialVLM~\cite{chen2024spatialvlm} conducts visual instruction tuning on metric depth-aware QA pairs for multi-modal large language models.
Different from these learning paradigms, we define 3D-awareness as the model's latent ability to estimate depth and pretrain an egocentric video-language model with a lightweight 3D-aware decoder.

\section{Method}
\label{sec:method}
Our objective is to develop a 3D-aware egocentric video-language model, where 3D-awareness is defined as the model's ability to infer depth information from its representations.
EgoDTM achieves this by pretraining depth-aware and text-aligned video representations.
~\Cref{fig:arch} illustrates the architecture of our proposed EgoDTM. 
We first describe the video-language model architecture in~\Cref{subsec:architecture}.  
Next, we introduce the 3D-aware video pretraining in~\Cref{subsec:3D-aware branch} and then present our approach to generate spatial-aware captions in~\Cref{subsec:spatial-aware text}. 
The details of training and inference are presented in~\Cref{subsec:training_and_inference}.

\subsection{Video-Language Model Architecture}
\label{subsec:architecture}

The video-language model basically consists of a video encoder and a text encoder. 

\noindent\textbf{Video and text encoders.}
Our video encoder employs a plain vision transformer~\cite{dosovitskiy2020image} backbone to process video tokens.
The video input $V\in\mathbb{R}^{F\times H\times W\times 3}$ is divided into $N=\frac{F}{F_0}\times\frac{H}{H_0}\times\frac{W}{W_0}$ non-overlapping cubes of dimension $F_0\times H_0\times W_0\times 3$, where $F$ is the number of frames, $H$ and $W$ denote height and width.
These cubes are combined with positional embeddings and processed by the video encoder to produce a feature map $\mathbf{v}_{Z}\in\mathbb{R}^{N\times C}$. 
Note that a ${\rm [CLS]}$ token is added to represent the global video embedding $\mathbf{v}_{\rm cls}$.  
For the text encoder, captions $T$ are tokenized via a Byte Pair Encoding (BPE) tokenizer~\cite{bpe} and encoded with a transformer initialized from CLIP~\cite{clip} to output the sentence embedding $\mathbf{t}_{\rm cls}$. 
Both encoders adopt the flash-attention~\cite{dao2022flashattention} to mitigate the memory bottleneck of the attention mechanism.

\noindent\textbf{Video-text alignment loss.}
We use the InfoNCE loss to align video and text embeddings.
For simplicity, we define the video-to-text loss within a batch $\mathcal{B}=\{(\mathbf{v}_i, \mathbf{t}_i\}_{i=1}^B$ as follows:
\begin{equation}
    \mathcal{L}_{\rm vtc}=-\frac{1}{B}\sum_{i} {\rm log}\frac{{\rm exp} (\mathbf{v}_i\cdot \mathbf{t}_i / \tau)}{\Sigma_{j} {\rm exp}(\mathbf{v}_i\cdot \mathbf{t}_j /\tau)}
\end{equation}
where $(\mathbf{v}_i, \mathbf{t}_i)$ represent the video and text embeddings $(\mathbf{v}_{\rm cls},\mathbf{t}_{\rm cls})$ for the $i$-th (video, caption) pair within the batch. The text-to-video loss is defined symmetrically.

\subsection{3D-Aware Video Pretraining}
A critical issue in leveraging depth maps for 3D-aware pretraining is that depth maps are typically high-resolution images (e.g., 1024p). However, most low-level details in these maps are redundant, as video-language models primarily benefit from the relative depth information between patches. Therefore, high-resolution depth maps may be unnecessary and inefficient for pretraining. Additionally, video-text pretraining often requires a large batch size to ensure good performance, further emphasizing the need for lightweight and memory-efficient processing. To address these concerns, we design a lightweight 3D-aware decoder to estimate low-resolution depth maps from video representations. Specifically, the decoder consists of a plain feature pyramid, a depth-aware transformer decoder, and two specialized heads, as described below.

\label{subsec:3D-aware branch}

\noindent\textbf{Plain feature pyramid.} 
Depth estimation methods~\cite{piccinelli2024unidepth,bhat2021adabins,ranftl2020midas} often utilize multi-scale features from intermediate backbone layers.
However, video-language pretraining typically uses only the final-layer representations. To reconcile these differences, we adopt a simplified FPN~\cite{li2022simplefpn} that projects the video feature map $\mathbf{v}_Z$ into multi-scale outputs $\{\mathbf{v}_{Z_{1\times}},\mathbf{v}_{Z_{2\times}},\cdots,\mathbf{v}_{Z_{m\times}}\}$,
where $\mathbf{v}_{Z_{m\times}}$ denotes the feature map of size $(N\times m^2)\times C$, and the size of the predicted depth maps equals $N\times m^2$.
Moreover, inspired by the adaptive bins~\cite{bhat2021adabins,bhat2022localbins,shao2023iebins,li2024binsformer} in depth estimation, we segment the depth range into intervals and predict distributions at each pixel over the interval centers.

\noindent\textbf{Depth-aware transformer decoder.} 
This module processes multi-scale features to generate depth-aware representations capable of estimating distinct distance ranges.
The transformer decoder follows standard architecture as~\cite{carion2020detr,cheng2022mask2former}, transforming $L$ learnable depth queries $\mathbf{Q}\in\mathbb{R}^{L\times C}$ into depth-aware representations $\mathbf{Z}\in\mathbb{R}^{L\times C}$ using self-attention and cross-attention mechanisms.
The cross attention processes $\mathbf{Q}$ as queries and concatenates $(\mathbf{v}_{Z_{1\times}},\mathbf{v}_{Z_{2\times}},\dots,\mathbf{v}_{Z_{m-1\times}})$ as key-value pairs $(k,v)$.
The learned depth queries estimate depths at different distances, as illustrated by varied gray colors in~\Cref{fig:arch}.

\noindent\textbf{Bin head. }
We partition the depth range of $[0,1]$ into $L$ intervals. 
The bin head $\mathcal{F}(\cdot):\mathbb{R}^C\rightarrow [0,1]$ processes depth-aware representations $\{\mathbf{z}_i\}_{i=1}^L$ to predict widths of depth interval bins $\{w_i\}_{i=0}^L$, satisfying $\sum_{i=0}^L w_i=1$ where we denote $w_0=0$.
The $i$-th depth interval is defined by boundaries $[\sum_{i=1}^{i-1}w_i,\sum_{i=1}^{i}w_i]$, with its center computed as:
\begin{equation}
    c_i = \sum_{i=1}^{i-1}w_i + \frac{w_i}{2}
\end{equation}

\noindent\textbf{Mask head.}
Then, we aim to predict distributions over the depth centers at each pixel.
Specifically, the mask head $\mathcal{G}(\cdot):\mathbb{R}^C\rightarrow \mathbb{R}^C$ takes $\mathbf{z}_i$ as input, and the output $\mathbf{m}_i$ is multiplied by the largest feature map $\mathbf{m}_i\otimes \mathbf{v}_{Z_{m\times}}$ to acquire unnormalized probabilistic maps $d_i\in\mathbb{R}^{N\times m^2}$, where $\otimes$ denotes the composition of element-wise multiplication and sum along the feature dimension.
After normalizing $\{d_i\}_{i=1}^L$ by the Softmax operation along the query dimension $L$, we get the probabilistic map $D=[D_1,\dots,D_L]\in\mathbb{R}^{L\times N\times m^2}$.
The process is described as:
\begin{equation}
    D=\mathrm{Softmax}(\mathcal{G}(\mathbf{Z})\otimes \mathbf{v}_{z_{m\times}})
\end{equation}
Finally, depth prediction $\tilde{D}\in\mathbb{R}^{N\times m^2}$ is obtained by the linear combination of the depth-bin-centers and the probabilistic maps:
{\setlength\abovedisplayskip{0.3cm}
\setlength\belowdisplayskip{0.3cm}
\begin{equation}
     \tilde{D} = \sum_{k=1}^L c_k D_k
\end{equation}
}

\noindent\textbf{3D-aware pretraining loss. }
Given the lack of depth annotations in large-scale egocentric videos~\cite{grauman2022ego4d}, we generate pseudo-depth labels using monocular depth estimation foundation models~\cite{ranftl2020midas,depthanything,depthanythingv2}.
The pretraining loss is defined as:
\begin{equation}
    \mathcal{L}_{\rm depth}=\|\tilde{D} - D_{{\rm gt}}\|_2
\end{equation}
where $\tilde{D}$ and $D_{\rm gt}$ are the predicted and ground-truths inverse depth maps within the range $[0, 1]$.

\begin{figure*}[t]
  \centering    \includegraphics[width=\linewidth]{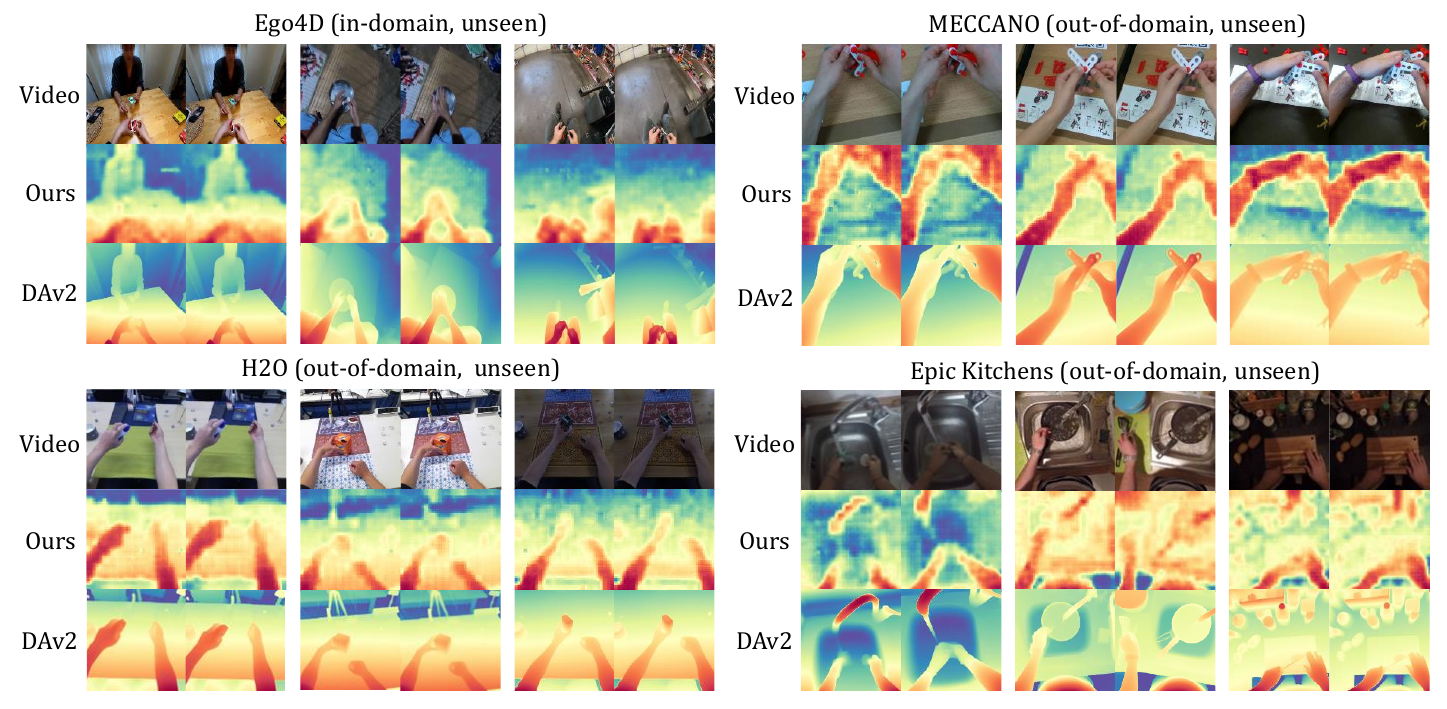}
  \caption{Qualitative results of depth estimation from EgoDTM and DepthAnythingv2~\cite{depthanythingv2}(DAv2) on datasets including in-domain but unseen Ego4D validation set~\cite{grauman2022ego4d}, out-of-domain and unseen data of EK100~\cite{epic-kitchens-100}, MECCANO~\cite{ragusa2021meccano}, and H2O~\cite{Kwon2021h2o}. 
  Note that DAv2 operates with a high resolution of 512p, while EgoDTM uses a lower resolution input of 224p and generates a depth map at a resolution of 56p. 
  Despite the lower resolution input, EgoDTM demonstrates intuitive generalization across unseen egocentric datasets with diverse environments, illuminations, backgrounds, and varying HOI object sizes.}
  \label{fig:vis_depth_estim}
\end{figure*}

\subsection{Spatial-Aware Textual Enrichment}
\label{subsec:spatial-aware text}

Previous egocentric video-language pretraining frameworks typically learn from short descriptions composed of verbs and nouns.
To foster 3D-aware pretraining, we argue that captions should contain more spatial-temporal information like HOI positions, object shapes, and movements.
We therefore enrich the captions with visual cues by employing a detect-track-generate pipeline tailored for egocentric videos using foundation models~\cite{hoidetector,sam2,deepseekllm}.

\noindent\textbf{Detect.} We use the detector~\cite{hoidetector}, HOID, finetuned on egocentric data to detect hands, objects, and their contact states for a given frame.
The detector $f(\cdot)$ receives $V_i$ as input and outputs diverse information $\{(b_j,c_j,t,s)\}_{j}$, where $b_j$ is the bounding boxes with confidence scores $c_j$, $t$ is object type, and $s$ is contact state of hand.

\noindent\textbf{Track.} Since predictions of HOID lack temporal consistency, we address this issue by tracking across the video. 
We utilize SAM2~\cite{sam2} $g(\cdot)$, a foundation model for promptable segmentation and tracking.
The initial tracking frame $V^*$ is selected when the hands make contact with objects.
The HOI masks of this frame are obtained by prompting SAM2 with HOI boxes $\{m^*_j\}_j=g(V^*,\{b^*_j\}_j)$, where the bounding boxes serve as negative prompts for each other.
Finally, the model tracks HOI masks throughout all frames to acquire the temporal consistent masks $M=(\{m_{1j}\}_{j},\dots,\{m_{Fj}\}_{j})$. 

\noindent\textbf{Generate.}
As suggested by previous works~\cite{liu2023llava,feng2023layoutgpt}, LLMs are capable of understanding visual cues like bounding boxes. 
We use an LLM $h(\cdot)$ to comprehend the HOI spatial information and original captions, then generate spatial-aware captions $T^{\text{aug}}=h(T, M)$.

Equipped with the enriched video-text pairs, our new video-text contrastive learning objective is:
\begin{equation}
    \mathcal{L}_{\rm augvtc}=-\frac{1}{B}\sum_{i} {\rm log}\frac{{\rm exp} (\mathbf{v}_i\cdot \mathbf{t}_i') / \tau)}{\Sigma_{j} {\rm exp}(\mathbf{v}_i\cdot \mathbf{t}_j' /\tau)}
\end{equation}
where $\mathbf{t}_i'={\rm rand}(\mathbf{t}_i, \mathbf{t}_i^{\rm aug})$ is randomly sampled from the original and enriched text embedding $\mathbf{t}_i^{\rm aug}$.

\subsection{Training Strategy}
\label{subsec:training_and_inference}

\noindent\textbf{Training and evaluation.}
We employ both the augmented video-text alignment loss and the 3D-aware pretraining loss to pretrain our model: 
\begin{equation}
\mathcal{L}=\mathcal{L}_{\rm augvtc}+\mathcal{L}_{\rm depth}
\end{equation}
Our pretraining improves video-language representations for downstream egocentric tasks without modifying the dual-encoders themselves. As a result, only the dual-encoders are used for downstream tasks, introducing no additional computational costs at inference time.

\noindent\textbf{Pretraining data.}
Our pretraining data consists of four million (video, text) pairs, with each video approximately 1 second long.
We generate enhanced text descriptions and depth maps specifically for videos featuring hand-object interactions, resulting in two million (video, enriched texts, depths) triplets.
Specifically, for each video, we extract eight depth maps using the DAv2-Large~\cite{depthanythingv2} model.
The HOI boxes are detected by HOID~\cite{hoidetector}, which is finetuned on egocentric data.
The HOI masks are segmented and tracked using the same unified SAM2-Large model~\cite{sam2}.
The final enriched text descriptions are generated using the DeepSeek-LLM-200B model~\cite{deepseekllm}.

\begin{table*}
\centering
\caption{Comparison with state-of-the-art methods on zero-shot video-text retrieval and action recognition.
All methods use the same 4M videos from Ego4D for pretraining and have similar hidden dimension sizes. 
The numbers of the method with $^*$ are sourced from~\cite{phan2024henasy}.
}
\setlength{\belowcaptionskip}{3pt}%
\scalebox{0.85}{
\begin{tabular}{l|cccc@{}|cccc@{}|ccc@{}|cc}
\toprule
& \multicolumn{7}{c}{\centering Epic-Kitchens-100-MIR} && \multicolumn{2}{c}{\centering EGTEA} && \multicolumn{2}{c}{\centering EgoMCQ}\\
\midrule
\multirow{2}{*}{\centering Method} &
\multicolumn{3}{c}{mAP (\%)} &&\multicolumn{3}{c}{nDCG (\%)} && \multirow{2}{*}{\centering mean} & \multirow{2}{*}{\centering top1} && \multirow{2}{*}{\centering Inter} & \multirow{2}{*}{\centering Intra}\\
& V$\rightarrow$ T & T$\rightarrow$ V & Avg. && V$\rightarrow$ T & T$\rightarrow$ V & Avg. &&& &&&\\
\midrule
EgoVLP~\cite{lin2022egovlp} & 
26.0 & 20.6 & 23.3 &&
28.8 & 27.0 & 27.9 &&
- & - &&
90.6 & 57.2 \\  
EgoVLPv2~\cite{pramanick2023egovlpv2} & 
35.1 & 26.6 & 30.8 &&
33.7 & 30.4 & 32.0 &&
30.9 & 35.1 &&
91.0 & 60.9 \\  
LaViLa~\cite{lavila} & 
35.1 & 26.6 & 30.8 &&
33.7 & 30.4 & 32.0 &&
30.9 & 35.1 &&
93.6 & 59.1 \\  
AVION~\cite{zhao2023avion} & 
{37.1} & {28.7} & {32.9} &&
34.4 & 31.0 & 32.7 &&
{38.6} & {42.3} &&
{94.4} & {62.1} \\
HelpingHands$^*$~\cite{zhang2023helping}  & 
35.6 & 26.8 & 31.2 &&
{34.7} & {31.7} & {33.2} &&
29.4 & 35.3 &&
93.2 & 58.8 \\
HENASY$^*$~\cite{phan2024henasy} & 
35.5 & 27.1 & 31.3 &&
34.6 & {31.7} & {33.2} &&
29.6 & 35.9 &&
94.1 & 61.3 \\
\midrule

EgoDTM (ours) & 
\textbf{37.9} & \textbf{29.1} & \textbf{33.5} &&
\textbf{34.8} & \textbf{31.9} & \textbf{33.4} &&
\textbf{40.2} & \textbf{43.2} &&
\textbf{94.6} &  \textbf{63.6} \\
\bottomrule
\end{tabular}}
\label{tab:comp_1_zeroshot_vtr_ar}
\end{table*}

\section{Experiments}
\label{sec:exp}
To evaluate our EgoDTM, we conduct experiments from three perspectives: short video understanding (video-text retrieval and action recognition), 3D space comprehension (depth estimation and robot manipulation), and long video understanding (natural language query and moment query).
These evaluations cover seven benchmarks across five datasets.
In the following sections, we detail the experimental setups (\Cref{subsec: experiment_setup}), main results (\Cref{subsec:main_results}), ablation studies (\Cref{subsec: ablation}), and further analyses (\Cref{subsec: analyses}).

\subsection{Benchmarks and Settings}
\label{subsec: experiment_setup}

\noindent\textbf{Zero-shot video-text retrieval (ZS-VTR).}
To assess video-text alignment, we conduct zero-shot retrieval evaluations on text-to-video multi-choice retrieval on Ego4D~\cite{grauman2022ego4d} (EgoMCQ) and multi-instance video-text retrieval on Epic-Kitchens-100~\cite{epic-kitchens-100} (EK100MIR). 
Following~\cite{lin2022egovlp,epic-kitchens-100}, we use mean Average Precision (mAP) and the normalized Discounted Cumulative Gain (nDCG) as metrics for EK100MIR and accuracy for EgoMCQ.

\noindent\textbf{Zero-shot action recognition (ZS-AR).}
Action recognition is conducted in a video-to-text retrieval manner.
We experiment on the EGTEA~\cite{egtea} and Epic-Kitchens-100~\cite{epic-kitchens-100} (EK100CLS). 
The task on EGTEA requires models to recognize 106 classes of cooking activities, 
and the EK100CLS includes evaluation on 97 verbs and 300 nouns in kitchens.
The metrics include mean accuracy, top-1 and top-5 accuracy across test splits.

\noindent\textbf{Depth estimation (DE).}
To assess the model's 3D awareness, we conduct depth estimation by finetuning our model on the H2O~\cite{Kwon2021h2o} dataset.
Following~\cite{cvpr-2024probing3D}, we add a prediction head composed of a linear layer and an upsampling convolution layer on top of the frozen video encoder.
Similar to~\cite{cvpr-2024probing3D,eigen2014depth}, the metrics are threshold accuracy ($\delta_i$): percentage of pixels $\tilde{D}_j$ in ground-truth depth maps that satisfy $\text{max}(\frac{D_{{\rm gt}j}}{\tilde{D}_j},\frac{\tilde{D}_j}{D_{{\rm gt}\ j}}) = \delta_i < thr^i$ for $thr=1.25,i\in\{1,2,3\}$, and RMSE: $\|\tilde{D} - {D}_{\rm gt}\|_2$. The scale-invariant metrics are shifted and normalized from scale-aware metrics. 


\begin{table}[tbp]
\caption{Comparisons of depth estimation task on H2O dataset.}
\centering
\scalebox{0.85}{

\begin{tabular}{l|cccc|cccc}
\toprule
\multirow{2}{*}{Method} & \multicolumn{4}{c}{Scale-Aware Metrics} & \multicolumn{4}{c}{Scale-Invariant Metrics} \\
 & $\delta_1$↑ & $\delta_2$↑ & $\delta_3$↑ & RMSE↓ & $\delta_1$↑ & $\delta_2$↑ & $\delta_3$↑ & RMSE↓ \\
\midrule
ConvNext~\cite{liu2022convnext} & 0.721 & 0.965 & 0.991 & 0.644 & 0.727 & 0.969 & 0.996 & 0.593 \\
CLIP~\cite{clip} & 0.795 & \textbf{0.966} & 0.988 & 0.624 & 0.811 & 0.976 & 0.994 & 0.559 \\
EgoVLP~\cite{lin2022egovlp} & 0.778 & 0.954 & 0.989 & 0.610 & \textbf{0.853} & \textbf{0.977} & 0.996 & 0.497 \\
LaViLa~\cite{lavila} & 0.801 & 0.954 & 0.987 & 0.598 & 0.811 & 0.964 & 0.993 & 0.552 \\
AVION~\cite{zhao2023avion} & 0.786 & 0.960 & 0.991 & 0.606 & 0.812 & 0.969 & 0.996 & 0.543 \\
\midrule
EgoDTM (ours)  & \textbf{0.826} & 0.964 & \textbf{0.993} & \textbf{0.539} & 0.848 & \textbf{0.977} & \textbf{0.998} & \textbf{0.481} \\
\bottomrule
\end{tabular}
}
\label{tab:comp_depth_estim_h2o}
\end{table}

\noindent\textbf{Robot Manipulation (RM).}
The robot must learn to accurately perceive and interact in 3D spaces to accomplish daily tasks. 
Following previous works~\cite{nair2022r3m,zeng2024mpi}, we evaluate the visual representations as frozen perception modules for downstream policy learning within the Franka Kitchens simulation environment~\cite{gupta2019frankakitchen}.
Five tasks are adopted: turn knob (TK), open door (OD), flip switch (FS), open microwave (OM), and slide door (SD).
We use success rate as the metric.

\noindent\textbf{Natural language queries (NLQ).}
The NLQ task localizes time intervals in a long video given a language query.
We experiment on the EgoNLQ task on Ego4D~\cite{grauman2022ego4d}.
To fairly evaluate different pretrained visual representations, we extract the visual features by video encoders and text features by the same pretrained BERT~\cite{devlin2019bert}, then train a VSLNet~\cite{zhang2020vslnet} to solve this task.
The evaluation metrics are “Rn@m”, where $n\in\{1,5\}$ and $m\in\{0.3, 0.5\}$, presenting the percentage of at least one of the top-$n$ predicted intervals having IoU greater than $m$.

\noindent\textbf{Moment Queries (MQ).}
The MQ task is a video-only problem that aims to detect all temporal activity intervals in a long video given a specified activity category.
We conduct experiments on the EgoMQ benchmark from Ego4D~\cite{grauman2022ego4d}.
Following the setup in~\cite{lin2022egovlp}, we extract visual features using video encoders and then train a VSGN model~\cite{zhao2021vsgn} to perform this task.
The evaluation metrics are “R@n, m”, consistent with those used in the NLQ task.

\noindent\textbf{Pretraining details.}
The dual-encoders are initialized by the checkpoint pretrained on the original four million video-text pairs~\cite{lin2022egovlp} from Ego4D.
EgoDTM is then trained for two epochs on 8*A800 GPUs, which requires approximately 10 hours and a learning rate of 3e-5.
The depth-aware transformer decoder comprises six layers. 
The hidden dimension of the dual encoders is 768, while the 3D-aware decoder uses a dimension of 256 for efficient design.
We use frames with 224p as input and 56p as output of the depth maps.
Consequently, our 3D-aware decoder only has 9M parameters, and the batch size is set to 4096.

\subsection{Main Results}
\label{subsec:main_results}

\noindent\textbf{Zero-shot video-text retrieval.}
In~\Cref{tab:comp_1_zeroshot_vtr_ar}, EgoDTM outperforms models like LaViLa~\cite{lavila} and AVION~\cite{zhao2023avion}. 
For example, on EK100-MIR, EgoDTM outperforms the state-of-the-art AVION by +0.9\% in mAP and +0.3\% in nDCG, while also achieving a +1.5\% improvement in intra accuracy on EgoMCQ.
This indicates that captions generated based on visual cues from our data curation pipeline are more accurate and informative than captions generated from visual-conditioned LLMs or simply text rewriting. 
Besides, our model surpasses HelpingHands~\cite{zhang2023helping} and HENASY~\cite{phan2024henasy}, which rely on noisy HOI detection supervision with extra parameters in the visual encoder, indicating that depth modality can potentially offer larger merits for video-language models.

\noindent\textbf{Zero-shot action recognition}
~\Cref{tab:comp_1_zeroshot_vtr_ar} illustrates the superior performance of EgoDTM on EGTEA.
While previous works have demonstrated that depth modality enhances traditional action recognition tasks~\cite{zhang2016rgbd,liu2019ntu}, their applicability has been largely constrained to specific scenarios and closed-set datasets.
Our results extend this understanding, showing that video-language models can effectively leverage depth modality to learn more generalizable video representations. 

\noindent\textbf{Depth estimation.}
We evaluate the ability of visual-language models to infer depth from images in~\Cref{tab:comp_depth_estim_h2o}.
Our model surpasses previous video-language models across most metrics by a large margin.
In particular, our method improves the scale-aware RMSE by approximately 9.8\% over the second-best result achieved by LaViLa.
Interestingly, EgoVLP performs better in scale-invariant $\delta_1$ metric, suggesting that it latently captures finer-grained pixel-level details.
We hypothesize that this advantage stems from EgoVLP's scene-aware negative sampling strategy, which samples video-text pairs captured from the same environments into the same batch, thereby learning relevant details.

\begin{table*}[t]
\setlength{\tabcolsep}{5pt} 
\begin{floatrow}
\capbtabbox{
\scalebox{0.66}{
\begin{tabular}{l|cccc}
\toprule
Method & R1@0.3 & R5@0.3 & R1@0.5 & R5@0.5 \\
\midrule
EgoVLP~\cite{lin2022egovlp} & 6.32 & 13.84 & 3.41 & 8.80 \\
LaViLa~\cite{lavila} & 7.12 & 14.82 & 3.87 & 9.55 \\
AVION~\cite{zhao2023avion}& 7.33 & 14.89 & 4.31 & 9.53   \\
\midrule
EgoDTM (ours) & \textbf{8.13} & \textbf{16.11} & \textbf{4.83} & \textbf{10.30} \\

\bottomrule
\end{tabular}
}
}{
\caption{Comparisons of NLQ task.}
\label{tab:comp_NLQ}
}

\capbtabbox{
\scalebox{0.66}{
\begin{tabular}{l|cccc}
\toprule
Method & R1@0.3 & R5@0.3 & R1@0.5 & R5@0.5 \\
\midrule
EgoVLP~\cite{lin2022egovlp} & 30.44 & 46.66 & 22.41 & 35.75 \\
LaViLa~\cite{lavila} & 32.9 & 48.68 & \textbf{24.12} & 37.59 \\
AVION~\cite{zhao2023avion}& 32.17 & 47.3 & 23.11 & 36.3   \\
\midrule
EgoDTM (ours) & \textbf{32.92} & \textbf{50.08} & 23.94 & \textbf{39.15} \\

\bottomrule
\end{tabular}
}
}{
 \caption{Comparisons of MQ task.}
 \label{tab:comp_MQ}
 \small
}

\end{floatrow}
\end{table*}





\begin{wraptable}{r}{0.5\textwidth}
\caption{Comparison of robot manipulation tasks in Franka Kitchen simulation to assess model's 3D-awareness.
}
\centering
\setlength{\belowcaptionskip}{3pt}%
\scalebox{0.59}{
\begin{tabular}{l|ccccc|c}
\toprule
Method & TK & OD & OM & FS & SD & Average\\
\midrule
\textcolor{gray}{R3M~\cite{nair2022r3m}} &
\textcolor{gray}{53.3\%} & \textcolor{gray}{50.7\%} & \textcolor{gray}{59.3\%} & \textcolor{gray}{86.3\%} & \textcolor{gray}{97.7\%} & \textcolor{gray}{69.4\%} \\
\textcolor{gray}{MPI~\cite{zeng2024mpi} }
& \textcolor{gray}{83.3\%} & \textcolor{gray}{54\%} & \textcolor{gray}{44.5\%} & \textcolor{gray}{93.5\%} & \textcolor{gray}{100\%} & \textcolor{gray}{75\%} \\
\midrule
ResNet~\cite{he2016resnet} & 28\% & 18\% & 26.7\% & 50\%   & 75.5\% & 39.7\% \\
CLIP~\cite{clip} & 26.3\% & 13\% & 24.7\% & 41.7\% & 86.3\% & 38.4\% \\
LaViLa~\cite{lavila} & 48\% & 26\% & 22.5\% & 69\% & \textbf{94.5\%} & 52\% \\
\midrule
EgoDTM (ours) & \textbf{56\%} & \textbf{28\%}& \textbf{35.5\%} & \textbf{81\%} & 92.5\% & \textbf{58.6\%}  \\
\bottomrule
\end{tabular}}
\label{tab:comp_3_RobotManip}
\end{wraptable}


\noindent\textbf{Robot manipulation.}
As shown in~\Cref{tab:comp_3_RobotManip}, EgoDTM consistently outperforms pretrained visual-language models such as CLIP~\cite{clip} and LaViLa~\cite{lavila} by +20.2\% and +6.6\%, respectively, demonstrating stronger spatial perception in visual representations.
Additionally, EgoDTM performs competitively with specialized robot learning methods on certain tasks, such as ``turn knob'' but underperforms on others, like ``open microwave''.
A possible reason is that our model, pretained on depth and text, may overfit to real-worold scenarios, whereas methods like R3M~\cite{nair2022r3m} leverage self-supervised pretraining, which could provide better generalization to diverse manipulation tasks.

\noindent\textbf{Natural language queries.}
As shown in ~\Cref{tab:comp_NLQ}, EgoDTM outperforms other egocentric video-language models, e.g., +1.22 on R5@0.3 over AVION.
In long video localization tasks, depth awareness enhances spatial understanding, 
enabling EgoDTM to better comprehend visual content.

\noindent\textbf{Moment queries.}
As shown in~\Cref{tab:comp_MQ}, EgoDTM achieves competitive performance among all methods. Specifically, it surpasses AVION by +2.78 on R5@0.3 and +2.85 on R5@0.5, and slightly improves over LaViLa by +1.4 on R5@0.3 and +1.56 on R5@0.5. These results demonstrate that depth-aware representation learning in EgoDTM effectively enhances spatial reasoning and temporal localization in long egocentric videos, leading to more accurate moment predictions.

\begin{table*}[t]
\caption{Zero-shot ablation studies. Models are pretrained on 4M video-text pairs from Ego4D and evaluated with zero-shot video-text retrieval on Epic-Kitchens-100-MIR~\cite{epic-kitchens-100} and EgoMCQ~\cite{lin2022egovlp}. Unless specified, the default setting includes: ViT-B/16 backbone, batch size of 2048, random substitution of enriched texts, both $\mathcal{L}_{\rm depth}$ and $\mathcal{L}_{\rm augvtc}$ as losses, 8 depth queries, and 56p depth maps.}
\centering
\begin{minipage}{0.35\textwidth}
\centering
\scalebox{0.6}{
\begin{tabular}{l|cc|cc}
\toprule
 & \multicolumn{2}{c}{\centering EK100MIR} & \multicolumn{2}{c}{\centering EgoMCQ}\\
& mAP & nDCG & Inter & Intra \\
\midrule
$\mathcal{L}_{\rm vtc}$  & 29.7 & 30.7 & 94.2 & 60.2 \\
$\mathcal{L}_{\rm vtc} + \mathcal{L}_{\rm depth}$ & 31.3 & 31.2 & 94.2 & 62.6 \\
$\mathcal{L}_{\rm augvtc}$ & 31.3 & 32.2 & 94.0 & 61.6 \\
$\mathcal{L}_{\rm augvtc} + \mathcal{L}_{\rm depth}$ & 33.1 & 33.1 & 94.6 & 62.6 \\
\bottomrule
\end{tabular}
}
\subcaption{\textbf{EgoDTM's components.} Both the 3D-aware video pretraining and textual enhancement enhance video-language representations.}
\label{tab:abl_1_components}
\end{minipage}
\hfill
\begin{minipage}{0.3\textwidth}
\centering
\scalebox{0.61}{
\begin{tabular}{c|cccc}
\toprule
  & \multicolumn{2}{c}{\centering EK100MIR} & \multicolumn{2}{c}{\centering EgoMCQ}\\
& mAP & nDCG & Inter & Intra \\
\midrule
4  & 32.6  & 32.5 & 94.61 & 62.61 \\
8  & 33.1  & 33.1 & 94.6 & 62.64 \\
16 & 32.3  & 32.4 & 94.72 & 62.46 \\
\bottomrule
\end{tabular}
} 
\label{tab:NumerOfQueries}
\subcaption{\textbf{Depth query numbers}. We set 8 queries as the default, which empirically generalizes best on EK100MIR.}
\end{minipage}
\hfill
\begin{minipage}{0.3\textwidth}
\centering
\scalebox{0.58}{
\begin{tabular}{cc|cccc}
\toprule
 \multirow{2}{*}{\centering Reso} & \multirow{2}{*}{\centering Mem} & \multicolumn{2}{c}{\centering EK100MIR} & \multicolumn{2}{c}{\centering EgoMCQ}\\
&  & mAP & nDCG & Inter & Intra \\
\midrule
28p   & 10G & 30.4    & 31.2    & 94.54 & 60.65\\
56p   & 17G & 30.9    & 31.9    & 94.28 & 61.05\\
112p  & 59G & 31.2 & 31.9 & 94.37 & 60.38 \\
\bottomrule
\end{tabular}
}
\label{tab:abl_6}
\subcaption{\textbf{Resolution of depth.} The batch size is set to 512 to accommodate the high GPU memory demands of large depth maps.}
\end{minipage}
\begin{minipage}{0.3\textwidth}
\centering
\scalebox{0.7}{
\begin{tabular}{c|cccc}
\toprule
  & \multicolumn{2}{c}{\centering EK100MIR} & \multicolumn{2}{c}{\centering EgoMCQ}\\
& mAP & nDCG & Inter & Intra \\
\midrule
Base & 33.1 & 33.1 & 94.6 & 62.64 \\
Large & 36.4 & 34.2 & 95.13 & 66.04 \\
\bottomrule
\end{tabular}
}
\label{tab:abl_4_ModelSize}
\subcaption{\textbf{Model size.} Large model brings more performance gain on intra accuracy of EgoMCQ.}
\end{minipage}
\hfill
\begin{minipage}{0.3\textwidth}
\centering
\scalebox{0.7}{
\begin{tabular}{c|cccc}
\toprule
  & \multicolumn{2}{c}{\centering EK100MIR} & \multicolumn{2}{c}{\centering EgoMCQ}\\
& mAP & nDCG & Inter & Intra \\
\midrule
All  & 32.8 & 32.7 & 94.46 & 61.92 \\
Rand   & 33.1 & 33.1 & 94.6 & 62.64 \\
\bottomrule
\end{tabular}
}
\label{tab: AugTextStrategy}
\subcaption{\textbf{Text Sampling Strategy.} Random substitution is better than thorough replacement.}
\end{minipage}
\hfill
\begin{minipage}{0.3\textwidth}
\centering
\scalebox{0.7}{
\begin{tabular}{c|cccc}
\toprule
  & \multicolumn{2}{c}{\centering EK100MIR} & \multicolumn{2}{c}{\centering EgoMCQ}\\
& mAP & nDCG & Inter & Intra \\
\midrule
1024 & 32.4 & 32.5 & 94.55 & 61.48 \\
2048 & 33.1 & 33.1 & 94.6 & 62.64 \\
4096 & 33.5 & 33.4 & 94.54 & 63.55 \\
\bottomrule
\end{tabular}
}
\subcaption{\textbf{Batch Size.} Increasing the batch size steadily yields larger improvements. }
\label{tab:abl_5_BatchSize}
\end{minipage}

\label{tab:ablations}
\end{table*}

\subsection{Ablation Study}
\label{subsec: ablation}

\begin{table*}[t]
\centering
\caption{Ablation studies on downstream tasks.}
\label{tab:ablation_more_tasks}
\resizebox{\textwidth}{!}{%
\begin{tabular}{lcccccc}
\toprule
\multirow{2}{*}{Metrics} & EK100MIR & EgoMCQ & EK100CLS & EgoNLQ & EgoMQ & DE \\
& mAP / nDCG $\uparrow$ & inter / intra acc $\uparrow$ & top-1 / top-5 acc $\uparrow$ & mIoU $\uparrow$ & mAP $\uparrow$ & scale-aware RMSE / scale-invariant RMSE $\downarrow$ \\
\midrule
$\mathcal{L}_{\text{vtc}}$ & 29.7 / 30.7 & 94.2 / 60.2 & 12.847 / 30.037 & 6.14 & 6.97 & 0.572 / 0.495 \\
$\mathcal{L}_{\text{vtc}} + \mathcal{L}_{\text{depth}}$ & 31.3 / 31.2 & 94.2 / \textbf{62.6} & 15.412 / 32.995 & 5.98 & 7.52 & 0.5637 / \textbf{0.464} \\
$\mathcal{L}_{\text{augvtc}}$ & 31.3 / 32.2 & 94 / 61.6 & \textbf{16.508} / 32.851 & \textbf{6.53} & 6.14 & 0.550 / 0.489 \\
$\mathcal{L}_{\text{augvtc}} + \mathcal{L}_{\text{depth}}$ & \textbf{33.1} / \textbf{33.1} & \textbf{94.6} / \textbf{62.6} & 15.898 / \textbf{33.895} & 6.17 & \textbf{8.87} & \textbf{0.539} / 0.481 \\
\bottomrule
\end{tabular}%
}
\label{tab: abl_loss}
\end{table*}
\noindent\textbf{EgoDTM's components.}
Combining video-text matching with depth estimation enhances each learning objective.
EgoDTM leverages depth maps to capture object relations, while textual descriptions enriched with shape and movement details can benefit more from depth information. 
As shown in the table~\Cref{tab: abl_loss}, progressively adding our modules leads to consistent improvements across the majority of tasks, and our model outperforms all baselines in the main experiments. One potential limitation is that depth pretraining may negatively impact the performance on EgoNLQ to some extent. Nevertheless, our proposed AugVTC module is able to mitigate this effect and yields improvements that surpass those achieved by Avion, LaViLa, and the EgoVLP encoder as shown in Table4 of our main paper. Besides, comparing row 2 and row 4 when adding $\mathcal{L}_{\text{augvtc}}$, we observe an increase in the scale-invariant RMSE and a decrease in the scale-aware RMSE. We hypothesize that this may result from the proxy task bias introduced by multi-task pretraining.

\begin{wraptable}{r}{0.45\linewidth}
\caption{Evaluation of our HOI mask generation pipeline on HOI segmentation benchmark VISOR~\cite{darkhalil2022visor} in kitchen environment.}
\centering
\scalebox{0.59}{
\begin{tabular}{c|c|cccc}
\toprule
Method & {\makecell[c]{mAP\\@(0.5:0.95)}} & {\makecell[c]{map\\@0.50}} & {\makecell[c]{hand-AP\\@(0.5:0.95)}} & {\makecell[c]{object-AP\\@(0.5:0.95)}} \\
\midrule

\textcolor{gray}{Upperbound~\cite{darkhalil2022visor}}  & \textcolor{gray}{60.70} & \textcolor{gray}{73.31} & \textcolor{gray}{90.87} & \textcolor{gray}{30.53} \\	
Our Pipeline & 43.02 & 54.88 & 61.01 & 25.03 \\
\bottomrule
\end{tabular}
}
\label{tab: analy_3_SegbyFM}
\end{wraptable}

\noindent\textbf{Depth query numbers.}
The number of depth queries affects depth granularity. 
We find that setting the query number to 8 is optimal, with each query covering a moderate depth range within $[0, 1]$.

\noindent\textbf{Resolution of the predicted depth map.}
The resolution of depth maps greatly impacts GPU memory usage. 
We find that higher-resolution depth maps slightly improve the video-text alignment, but this comes at the cost of reducing the maximum batch size.
Therefore, we choose a resolution to 56p as a balanced trade-off between high resolution and the ability to maintain a large batch size.

\noindent\textbf{Model size.}
A larger model size enhances performance, especially on the EgoMCQ intra task, which requires selecting the correct video from visually similar alternatives.

\noindent\textbf{Text sampling strategy.}
We examine the impact of text sampling strategy on model performance. A mixed sampling strategy that includes both enriched and original texts for pretraining yields stronger results by leveraging both detailed and simplified textual information.

\noindent\textbf{Batch size.}
The experiment results show that increasing batch size steadily improves performance.

\subsection{More Analysis}
\label{subsec: analyses}



\begin{figure*}[t]
\centering
\begin{floatrow}
\ffigbox[\FBwidth]{%
\caption{Comparisons of the noisy HOI bounding boxes (left) and the spatial-temporal consistent HOI masks (right).}%
\label{fig:Consistent_HOI_masks}%
}{%
\includegraphics[width=\linewidth]{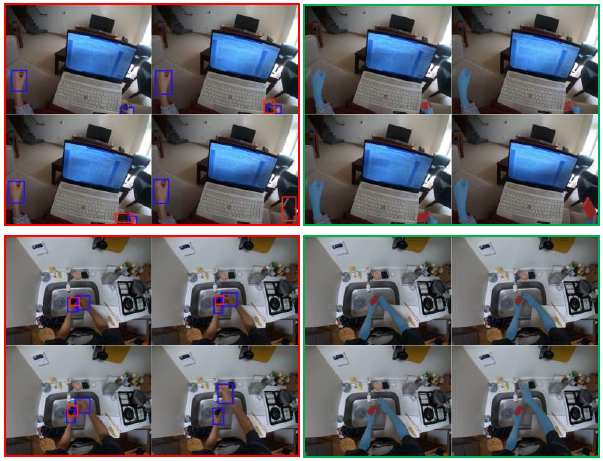}%
}
\ffigbox[\FBwidth]{%
\caption{Example of generalizable data construction. For better visualizations, we blur the background to highlight HOI regions.}%
\label{fig:case_data_construction}%
}{%
\includegraphics[width=\linewidth]{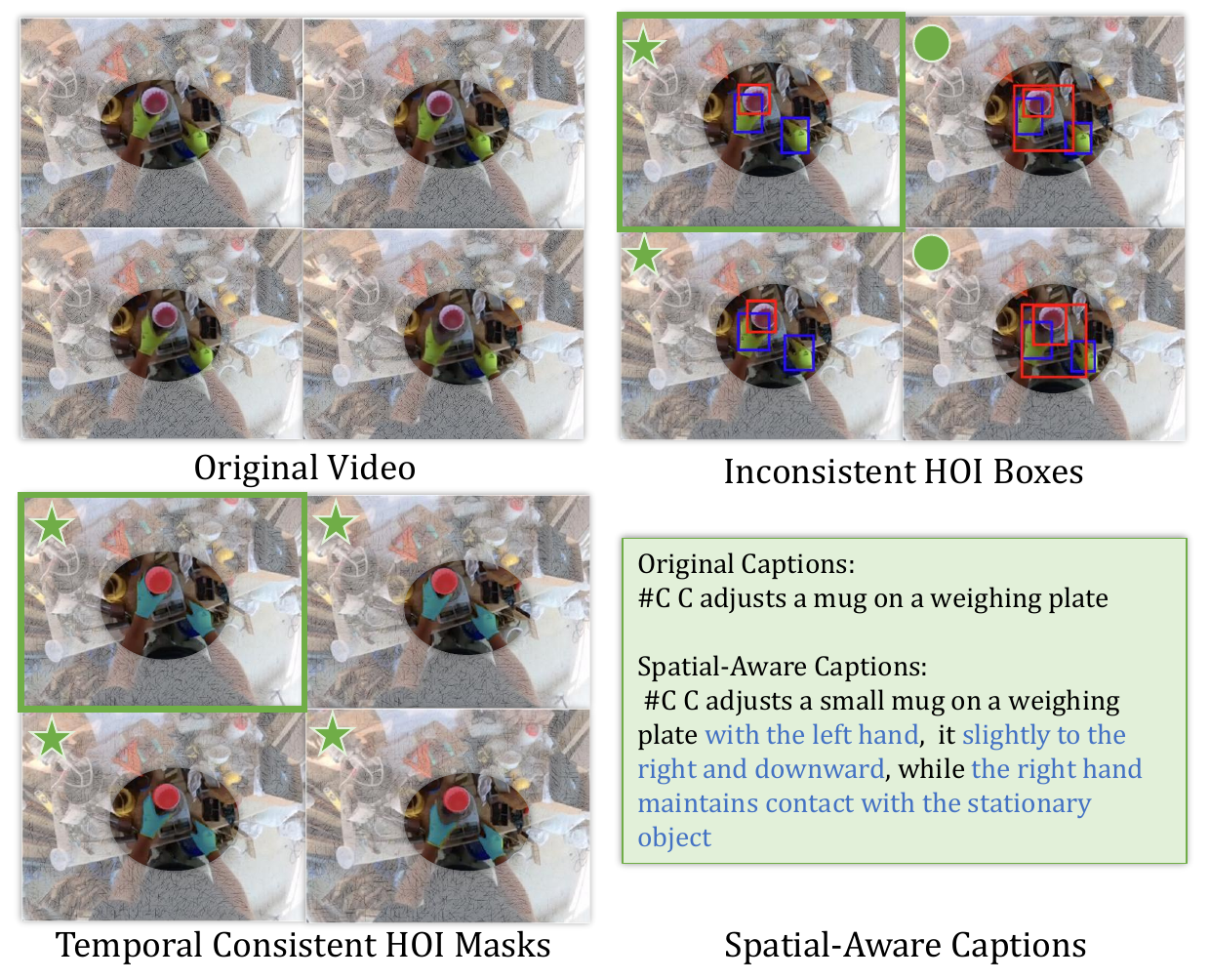}%
}
\end{floatrow}
\end{figure*}

\noindent\textbf{Quantitative analyses of the HOI mask generation pipeline.}
To evaluate the reliability of the HOI detector and SAM2 model in the egocentric domain, particularly for hand-object interactions, we evaluate the image-based HOI segmentation on the VISOR dataset~\cite{darkhalil2022visor}. The results are presented in~\Cref{tab: analy_3_SegbyFM}.
The supervised training model on VISOR serves as the upper bound for performance comparison.
While our generation pipeline does not fully reach the upper bound, it demonstrates high-quality segmentation results.
Notably, hand segmentation is significantly lower than the upper bound, likely due to inaccuracies in HOI detection prompts and challenges in segmenting hand-object interactions with SAM2 in cluttered egocentric backgrounds.

\noindent\textbf{Qualitative analyses of the HOI mask generation pipeline.}
We firstly compare inconsistent HOI bounding boxes from HOID~\cite{hoidetector} (frame-by-frame detection) with spatial-temporal HOI masks generated by combining HOID~\cite{hoidetector} and SAM2~\cite{sam2} in~\Cref{fig:Consistent_HOI_masks}.
The HOI boxes detected by HOI detector often lose tracks, since they are detected by an image-based model.
Our pipeline achieves consistent HOI tracking across frames, offering more precise HOI labels. 
~\Cref{fig:case_data_construction} presents that our data construction pipeline is useful even when HOI regions are small and the background is noisy.


\section{Conclusion}

In this work, we present EgoDTM, a novel egocentric 3D-aware video-language model. 
EgoDTM integrates dual transformer encoders with a lightweight 3D-aware depth decoder,  trained using video-text contrastive learning and depth estimation objectives.
To enable large-scale pretraining, we generate millions of depth maps and spatially enriched captions by leveraging foundation models. The captions are enhanced through a detect-track-generate pipeline specifically tailored for egocentric videos. 
EgoDTM demonstrates intuitive generalization in estimating depths in unseen environments.
Extensive experiments across diverse benchmarks, spanning short video understanding, 3D understanding, and long video understanding, validate the effectiveness of our approach.

\noindent\textbf{Discussions.}
While EgoDTM demonstrates strong performance in egocentric hand-object interaction scenarios, its generalization to broader indoor scenarios remains limited. 
Further exploration may include integrating 3D-aware visual encoders into multimodal large language models to enhance spatial awareness. 
Moreover, pretraining large-scale spatial-aware egocentric models with richer 3D signals, as explored in VGGT~\cite{wang2025vggt}, remains a promising yet challenging direction.

\clearpage
\section*{Acknowledgments}
This work was partially supported by the Beijing Natural Science Foundation (No. L233008) and the Outstanding Innovative Talents Cultivation Funded Programs 2024 of Renmin University of China.

{
    \small
    \bibliographystyle{plain}
    \bibliography{main}

@String(CVPR= {IEEE Conf. Comput. Vis. Pattern Recog.})

@String(ICCV= {Int. Conf. Comput. Vis.})

@String(ECCV= {Eur. Conf. Comput. Vis.})

@String(BMVC= {Brit. Mach. Vis. Conf.})

@String(CVPR  = {CVPR})

@String(ICCV  = {ICCV})

@String(ECCV  = {ECCV})

@String(BMVC  =	{BMVC})

@article{vaswani2017attention,
  title={Attention is all you need},
  author={Vaswani, Ashish and Shazeer, Noam and Parmar, Niki and Uszkoreit, Jakob and Jones, Llion and Gomez, Aidan N and Kaiser, {\L}ukasz and Polosukhin, Illia},
  journal={Advances in neural information processing systems},
  volume={30},
  year={2017}
}

@inproceedings{
dosovitskiy2020image,
title={An Image is Worth 16x16 Words: Transformers for Image Recognition at Scale},
author={Alexey Dosovitskiy and Lucas Beyer and Alexander Kolesnikov and Dirk Weissenborn and Xiaohua Zhai and Thomas Unterthiner and Mostafa Dehghani and Matthias Minderer and Georg Heigold and Sylvain Gelly and Jakob Uszkoreit and Neil Houlsby},
booktitle={International Conference on Learning Representations},
year={2021},
url={https://openreview.net/forum?id=YicbFdNTTy}
}

@inproceedings{cvpr-2024probing3D,
  title={Probing the 3d awareness of visual foundation models},
  author={El Banani, Mohamed and Raj, Amit and Maninis, Kevis-Kokitsi and Kar, Abhishek and Li, Yuanzhen and Rubinstein, Michael and Sun, Deqing and Guibas, Leonidas and Johnson, Justin and Jampani, Varun},
  booktitle={Proceedings of the IEEE/CVF Conference on Computer Vision and Pattern Recognition},
  pages={21795--21806},
  year={2024}
}

@inproceedings{nips-man2024lexicon3d,
      title={Lexicon3D: Probing Visual Foundation Models for Complex 3D Scene Understanding},
      author={Man, Yunze and Zheng, Shuhong and Bao, Zhipeng and Hebert, Martial and Gui, Liang-Yan and Wang, Yu-Xiong},
      booktitle={Advances in Neural Information Processing Systems},
      year={2024} 
      }

@article{lin2022egovlp,
  title={Egocentric video-language pretraining},
  author={Lin, Kevin Qinghong and Wang, Jinpeng and Soldan, Mattia and Wray, Michael and Yan, Rui and Xu, Eric Z and Gao, Difei and Tu, Rong-Cheng and Zhao, Wenzhe and Kong, Weijie and others},
  journal={Advances in Neural Information Processing Systems},
  volume={35},
  pages={7575--7586},
  year={2022}
}

@inproceedings{bpe,
    title = "Neural Machine Translation of Rare Words with Subword Units",
    author = "Sennrich, Rico  and
      Haddow, Barry  and
      Birch, Alexandra",
    booktitle = "Proceedings of the 54th Annual Meeting of the Association for Computational Linguistics (Volume 1: Long Papers)",
    month = aug,
    year = "2016",
    address = "Berlin, Germany",
    publisher = "Association for Computational Linguistics",
    url = "https://www.aclweb.org/anthology/P16-1162",
    doi = "10.18653/v1/P16-1162",
    pages = "1715--1725",
}

@inproceedings{
phan2024henasy,
title={HENASY: Learning to Assemble Scene-Entities for Interpretable Egocentric Video-Language Model},
author={Phan, Khoa Vo Thinh and Tran, Kashu Yamazaki Minh and Le, Ngan},
booktitle={Advances in Neural Information Processing Systems},
year={2024},
}

@inproceedings{bain2021frozen,
  title={Frozen in time: A joint video and image encoder for end-to-end retrieval},
  author={Bain, Max and Nagrani, Arsha and Varol, G{\"u}l and Zisserman, Andrew},
  booktitle={Proceedings of the IEEE/CVF International Conference on Computer Vision},
  pages={1728--1738},
  year={2021}
}

@inproceedings{pramanick2023egovlpv2,
  title={Egovlpv2: Egocentric video-language pre-training with fusion in the backbone},
  author={Pramanick, Shraman and Song, Yale and Nag, Sayan and Lin, Kevin Qinghong and Shah, Hardik and Shou, Mike Zheng and Chellappa, Rama and Zhang, Pengchuan},
  booktitle={Proceedings of the IEEE/CVF International Conference on Computer Vision},
  pages={5285--5297},
  year={2023}
}

@inproceedings{ashutosh2023hiervl,
  title={Hiervl: Learning hierarchical video-language embeddings},
  author={Ashutosh, Kumar and Girdhar, Rohit and Torresani, Lorenzo and Grauman, Kristen},
  booktitle={Proceedings of the IEEE/CVF Conference on Computer Vision and Pattern Recognition},
  pages={23066--23078},
  year={2023}
}

@inproceedings{clip,
  title={Learning transferable visual models from natural language supervision},
  author={Radford, Alec and Kim, Jong Wook and Hallacy, Chris and Ramesh, Aditya and Goh, Gabriel and Agarwal, Sandhini and Sastry, Girish and Askell, Amanda and Mishkin, Pamela and Clark, Jack and others},
  booktitle={International conference on machine learning},
  pages={8748--8763},
  year={2021},
  organization={PMLR}
}

@inproceedings{grauman2022ego4d,
  title={Ego4d: Around the world in 3,000 hours of egocentric video},
  author={Grauman, Kristen and Westbury, Andrew and Byrne, Eugene and Chavis, Zachary and Furnari, Antonino and Girdhar, Rohit and Hamburger, Jackson and Jiang, Hao and Liu, Miao and Liu, Xingyu and others},
  booktitle={Proceedings of the IEEE/CVF Conference on Computer Vision and Pattern Recognition},
  pages={18995--19012},
  year={2022}
}

@inproceedings{lavila,
  title={Learning video representations from large language models},
  author={Zhao, Yue and Misra, Ishan and Kr{\"a}henb{\"u}hl, Philipp and Girdhar, Rohit},
  booktitle={Proceedings of the IEEE/CVF Conference on Computer Vision and Pattern Recognition},
  pages={6586--6597},
  year={2023}
}

@article{luo2022clip4clip,
  title={Clip4clip: An empirical study of clip for end to end video clip retrieval and captioning},
  author={Luo, Huaishao and Ji, Lei and Zhong, Ming and Chen, Yang and Lei, Wen and Duan, Nan and Li, Tianrui},
  journal={Neurocomputing},
  volume={508},
  pages={293--304},
  year={2022},
  publisher={Elsevier}
}

@ARTICLE{epic-kitchens-100,
title={The EPIC-KITCHENS Dataset: Collection, Challenges and Baselines},
author={Damen, Dima and Doughty, Hazel and Farinella, Giovanni Maria  and Fidler, Sanja and 
       Furnari, Antonino and Kazakos, Evangelos and Moltisanti, Davide and Munro, Jonathan 
       and Perrett, Toby and Price, Will and Wray, Michael},
journal={IEEE Transactions on Pattern Analysis and Machine Intelligence (TPAMI)},
year={2021},
volume={43},
number={11},
pages={4125-4141},
doi={10.1109/TPAMI.2020.2991965}
}

@inproceedings{egtea,
  title={In the eye of beholder: Joint learning of gaze and actions in first person video},
  author={Li, Yin and Liu, Miao and Rehg, James M},
  booktitle={Proceedings of the European conference on computer vision (ECCV)},
  pages={619--635},
  year={2018}
}

@inproceedings{charades-ego,
  title={Actor and observer: Joint modeling of first and third-person videos},
  author={Sigurdsson, Gunnar A and Gupta, Abhinav and Schmid, Cordelia and Farhadi, Ali and Alahari, Karteek},
  booktitle={proceedings of the IEEE conference on computer vision and pattern recognition},
  pages={7396--7404},
  year={2018}
}

@article{openset-nips2023,
  title={Opening the vocabulary of egocentric actions},
  author={Chatterjee, Dibyadip and Sener, Fadime and Ma, Shugao and Yao, Angela},
  journal={Advances in Neural Information Processing Systems},
  volume={36},
  year={2024}
}

@article{gpt2,
  title={Language models are unsupervised multitask learners},
  author={Radford, Alec and Wu, Jeffrey and Child, Rewon and Luan, David and Amodei, Dario and Sutskever, Ilya and others},
  journal={OpenAI blog},
  volume={1},
  number={8},
  pages={9},
  year={2019}
}

@article{t5,
  title={Exploring the limits of transfer learning with a unified text-to-text transformer},
  author={Raffel, Colin and Shazeer, Noam and Roberts, Adam and Lee, Katherine and Narang, Sharan and Matena, Michael and Zhou, Yanqi and Li, Wei and Liu, Peter J},
  journal={Journal of machine learning research},
  volume={21},
  number={140},
  pages={1--67},
  year={2020}
}

@article{actor_observer,
  title={Actor and Observer: Joint Modeling of First and Third-Person Videos},
  author={Gunnar A. Sigurdsson and Abhinav Kumar Gupta and Cordelia Schmid and Ali Farhadi and Alahari Karteek},
  journal={2018 IEEE/CVF Conference on Computer Vision and Pattern Recognition},
  year={2018},
  pages={7396-7404},
  url={https://api.semanticscholar.org/CorpusID:4562167}
}

@inproceedings{xu2023pov,
  title={POV: Prompt-Oriented View-Agnostic Learning for Egocentric Hand-Object Interaction in the Multi-view World},
  author={Xu, Boshen and Zheng, Sipeng and Jin, Qin},
  booktitle={Proceedings of the 31st ACM International Conference on Multimedia},
  pages={2807--2816},
  year={2023}
}

@inproceedings{caba2015activitynet,
  title={Activitynet: A large-scale video benchmark for human activity understanding},
  author={Caba Heilbron, Fabian and Escorcia, Victor and Ghanem, Bernard and Carlos Niebles, Juan},
  booktitle={Proceedings of the ieee conference on computer vision and pattern recognition},
  pages={961--970},
  year={2015}
}

@article{mangalam2024egoschema,
  title={Egoschema: A diagnostic benchmark for very long-form video language understanding},
  author={Mangalam, Karttikeya and Akshulakov, Raiymbek and Malik, Jitendra},
  journal={Advances in Neural Information Processing Systems},
  volume={36},
  year={2024}
}

@inproceedings{zhang2023helping,
  title={Helping hands: An object-aware ego-centric video recognition model},
  author={Zhang, Chuhan and Gupta, Ankush and Zisserman, Andrew},
  booktitle={Proceedings of the IEEE/CVF International Conference on Computer Vision},
  pages={13901--13912},
  year={2023}
}

@article{plizzari2023outlook,
  title={An outlook into the future of egocentric vision},
  author={Plizzari, Chiara and Goletto, Gabriele and Furnari, Antonino and Bansal, Siddhant and Ragusa, Francesco and Farinella, Giovanni Maria and Damen, Dima and Tommasi, Tatiana},
  journal={International Journal of Computer Vision},
  pages={1--57},
  year={2024},
  publisher={Springer}
}

@inproceedings{xu2023egopca,
  title={EgoPCA: A New Framework for Egocentric Hand-Object Interaction Understanding},
  author={Xu, Yue and Li, Yong-Lu and Huang, Zhemin and Liu, Michael Xu and Lu, Cewu and Tai, Yu-Wing and Tang, Chi-Keung},
  booktitle={Proceedings of the IEEE/CVF International Conference on Computer Vision},
  pages={5273--5284},
  year={2023}
}

@inproceedings{plizzari2023can,
  title={What can a cook in Italy teach a mechanic in India? Action Recognition Generalisation Over Scenarios and Locations},
  author={Plizzari, Chiara and Perrett, Toby and Caputo, Barbara and Damen, Dima},
  booktitle={Proceedings of the IEEE/CVF International Conference on Computer Vision},
  pages={13656--13666},
  year={2023}
}

@inproceedings{li2021egoexo,
  title={Ego-exo: Transferring visual representations from third-person to first-person videos},
  author={Li, Yanghao and Nagarajan, Tushar and Xiong, Bo and Grauman, Kristen},
  booktitle={Proceedings of the IEEE/CVF Conference on Computer Vision and Pattern Recognition},
  pages={6943--6953},
  year={2021}
}

@INPROCEEDINGS{kazakos2021little,
  author={Kazakos, Evangelos and Huh, Jaesung and Nagrani, Arsha and Zisserman, Andrew and Damen, Dima},
  booktitle={British Machine Vision Conference (BMVC)},
  title={With a Little Help from my Temporal Context: Multimodal Egocentric Action Recognition},
  year={2021}}

@inproceedings{zhang2022fine,
  title={Fine-grained egocentric hand-object segmentation: Dataset, model, and applications},
  author={Zhang, Lingzhi and Zhou, Shenghao and Stent, Simon and Shi, Jianbo},
  booktitle={European Conference on Computer Vision},
  pages={127--145},
  year={2022},
  organization={Springer}
}

@article{Ryan2023EgocentricAA,
  title={Egocentric Auditory Attention Localization in Conversations},
  author={Fiona Ryan and Hao Jiang and Abhinav Shukla and James M. Rehg and Vamsi Krishna Ithapu},
  journal={2023 IEEE/CVF Conference on Computer Vision and Pattern Recognition (CVPR)},
  year={2023},
  pages={14663-14674},
  url={https://api.semanticscholar.org/CorpusID:257771767}
}

@inproceedings{hoidetector,
  title={Understanding human hands in contact at internet scale},
  author={Shan, Dandan and Geng, Jiaqi and Shu, Michelle and Fouhey, David F},
  booktitle={Proceedings of the IEEE/CVF conference on computer vision and pattern recognition},
  pages={9869--9878},
  year={2020}
}

@inproceedings{liu2022hoi4d,
  title={HOI4D: A 4D Egocentric Dataset for Category-Level Human-Object Interaction},
  author={Liu, Yunze and Liu, Yun and Jiang, Che and Lyu, Kangbo and Wan, Weikang and Shen, Hao and Liang, Boqiang and Fu, Zhoujie and Wang, He and Yi, Li},
  booktitle={Proceedings of the IEEE/CVF Conference on Computer Vision and Pattern Recognition},
  pages={21013--21022},
  year={2022}
}

@inproceedings{ragusa2021meccano,
  title = {The MECCANO Dataset: Understanding Human-Object Interactions from Egocentric Videos in an Industrial-like Domain},
  author = {Francesco Ragusa and Antonino Furnari and Salvatore Livatino and Giovanni Maria Farinella},
  year = {2021},
  eprint = {2010.05654},
  booktitle = {IEEE Winter Conference on Application of Computer Vision (WACV)}
}

@InProceedings{Kwon2021h2o,
author = {Kwon, Taein and Tekin, Bugra and St\"uhmer, Jan and Bogo, Federica and Pollefeys, Marc},
title = {H2O: Two Hands Manipulating Objects for First Person Interaction Recognition},
booktitle = {Proceedings of the IEEE/CVF International Conference on Computer Vision (ICCV)},
month = {October},
year = {2021},
pages = {10138-10148}
}

@inproceedings{depthanything,
      title={Depth Anything: Unleashing the Power of Large-Scale Unlabeled Data}, 
      author={Yang, Lihe and Kang, Bingyi and Huang, Zilong and Xu, Xiaogang and Feng, Jiashi and Zhao, Hengshuang},
      booktitle={CVPR},
      year={2024}
}

@misc{ramakrishnan2024doesspatialcognitionemerge,
      title={Does Spatial Cognition Emerge in Frontier Models?}, 
      author={Santhosh Kumar Ramakrishnan and Erik Wijmans and Philipp Kraehenbuehl and Vladlen Koltun},
      year={2024},
      eprint={2410.06468},
      archivePrefix={arXiv},
      primaryClass={cs.AI},
      url={https://arxiv.org/abs/2410.06468}, 
}

@inproceedings{tong2024eyes,
  title={Eyes wide shut? exploring the visual shortcomings of multimodal llms},
  author={Tong, Shengbang and Liu, Zhuang and Zhai, Yuexiang and Ma, Yi and LeCun, Yann and Xie, Saining},
  booktitle={Proceedings of the IEEE/CVF Conference on Computer Vision and Pattern Recognition},
  pages={9568--9578},
  year={2024}
}

@inproceedings{
nair2022r3m,
title={R3M: A Universal Visual Representation for Robot Manipulation},
author={Suraj Nair and Aravind Rajeswaran and Vikash Kumar and Chelsea Finn and Abhinav Gupta},
booktitle={6th Annual Conference on Robot Learning},
year={2022},
url={https://openreview.net/forum?id=tGbpgz6yOrI}
}

@inproceedings{zeng2024mpi,
    title={Learning Manipulation by Predicting Interaction},
    author={Jia, Zeng and Qingwen, Bu and Bangjun, Wang and Wenke, Xia and Li, Chen and Hao, Dong and Haoming, Song and Dong, Wang and Di, Hu and Ping, Luo and Heming, Cui and Bin, Zhao and Xuelong, Li and Yu, Qiao and Hongyang, Li},
    booktitle= {Proceedings of Robotics: Science and Systems (RSS)},
    year={2024}
  }

@inproceedings{li2022simplefpn,
  title={Exploring plain vision transformer backbones for object detection},
  author={Li, Yanghao and Mao, Hanzi and Girshick, Ross and He, Kaiming},
  booktitle={European conference on computer vision},
  pages={280--296},
  year={2022},
  organization={Springer}
}

@inproceedings{cheng2022mask2former,
  title={Masked-attention mask transformer for universal image segmentation},
  author={Cheng, Bowen and Misra, Ishan and Schwing, Alexander G and Kirillov, Alexander and Girdhar, Rohit},
  booktitle={Proceedings of the IEEE/CVF conference on computer vision and pattern recognition},
  pages={1290--1299},
  year={2022}
}

@article{ranftl2020midas,
  title={Towards robust monocular depth estimation: Mixing datasets for zero-shot cross-dataset transfer},
  author={Ranftl, Ren{\'e} and Lasinger, Katrin and Hafner, David and Schindler, Konrad and Koltun, Vladlen},
  journal={IEEE transactions on pattern analysis and machine intelligence},
  volume={44},
  number={3},
  pages={1623--1637},
  year={2020},
  publisher={IEEE}
}

@inproceedings{bhat2021adabins,
  title={Adabins: Depth estimation using adaptive bins},
  author={Bhat, Shariq Farooq and Alhashim, Ibraheem and Wonka, Peter},
  booktitle={Proceedings of the IEEE/CVF conference on computer vision and pattern recognition},
  pages={4009--4018},
  year={2021}
}

@inproceedings{dao2022flashattention,
  title={Flash{A}ttention: Fast and Memory-Efficient Exact Attention with {IO}-Awareness},
  author={Dao, Tri and Fu, Daniel Y. and Ermon, Stefano and Rudra, Atri and R{\'e}, Christopher},
  booktitle={Advances in Neural Information Processing Systems (NeurIPS)},
  year={2022}
}

@article{zhao2023avion,
  title={Training a large video model on a single machine in a day},
  author={Zhao, Yue and Kr{\"a}henb{\"u}hl, Philipp},
  journal={arXiv preprint arXiv:2309.16669},
  year={2023}
}

@inproceedings{depthanythingv2,
  title={Depth Anything V2},
  author={Yang, Lihe and Kang, Bingyi and Huang, Zilong and Zhao, Zhen and Xu, Xiaogang and Feng, Jiashi and Zhao, Hengshuang},
  booktitle={Advances in Neural Information Processing Systems (NeurIPS)},
  year={2024}
}

@article{gupta2019frankakitchen,
  title={Relay Policy Learning: Solving Long Horizon Tasks via Imitation and Reinforcement Learning},
  author={Gupta, Abhishek and Kumar, Vikash and Lynch, Corey and Levine, Sergey and Hausman, Karol},
  journal={Conference on Robot Learning (CoRL)},
  year={2019}
}

@article{darkhalil2022visor,
  title={Epic-kitchens visor benchmark: Video segmentations and object relations},
  author={Darkhalil, Ahmad and Shan, Dandan and Zhu, Bin and Ma, Jian and Kar, Amlan and Higgins, Richard and Fidler, Sanja and Fouhey, David and Damen, Dima},
  journal={Advances in Neural Information Processing Systems},
  volume={35},
  pages={13745--13758},
  year={2022}
}

@article{hu2023robotsurvey,
  title={Toward general-purpose robots via foundation models: A survey and meta-analysis},
  author={Hu, Yafei and Xie, Quanting and Jain, Vidhi and Francis, Jonathan and Patrikar, Jay and Keetha, Nikhil and Kim, Seungchan and Xie, Yaqi and Zhang, Tianyi and Fang, Hao-Shu and others},
  journal={arXiv preprint arXiv:2312.08782},
  year={2023}
}

@article{mccarthy2024survey2,
  title={Towards Generalist Robot Learning from Internet Video: A Survey},
  author={McCarthy, Robert and Tan, Daniel CH and Schmidt, Dominik and Acero, Fernando and Herr, Nathan and Du, Yilun and Thuruthel, Thomas G and Li, Zhibin},
  journal={arXiv preprint arXiv:2404.19664},
  year={2024}
}

@article{deepseekllm,
  author = {DeepSeek-AI},
  title = {DeepSeek LLM: Scaling Open-Source Language Models with Longtermism},
  journal = {arXiv preprint arXiv:2401.02954},
  year = {2024},
  url = {https://github.com/deepseek-ai/DeepSeek-LLM}
}

@article{dpt,
	author    = {Ren\'{e} Ranftl and Katrin Lasinger and David Hafner and Konrad Schindler and Vladlen Koltun},
	title     = {Towards Robust Monocular Depth Estimation: Mixing Datasets for Zero-shot Cross-dataset Transfer},
	journal   = {IEEE Transactions on Pattern Analysis and Machine Intelligence (TPAMI)},
	year      = {2020},
}

@inproceedings{he2016resnet,
  title={Deep residual learning for image recognition},
  author={He, Kaiming and Zhang, Xiangyu and Ren, Shaoqing and Sun, Jian},
  booktitle={Proceedings of the IEEE conference on computer vision and pattern recognition},
  pages={770--778},
  year={2016}
}

@inproceedings{yue2024fit3d,
  title     = {{Improving 2D Feature Representations by 3D-Aware Fine-Tuning}},
  author    = {Yue, Yuanwen and Das, Anurag and Engelmann, Francis and Tang, Siyu and Lenssen, Jan Eric},
  booktitle = {European Conference on Computer Vision (ECCV)},
  year      = {2024}
}

@article{koenderink1992surface,
  title={Surface shape and curvature scales},
  author={Koenderink, Jan J and Van Doorn, Andrea J},
  journal={Image and vision computing},
  volume={10},
  number={8},
  pages={557--564},
  year={1992},
  publisher={Elsevier}
}

@article{spelke2010beyond,
  title={Beyond core knowledge: Natural geometry},
  author={Spelke, Elizabeth and Lee, Sang Ah and Izard, V{\'e}ronique},
  journal={Cognitive science},
  volume={34},
  number={5},
  pages={863--884},
  year={2010},
  publisher={Wiley Online Library}
}

@article{shepard1971mental,
  title={Mental rotation of three-dimensional objects},
  author={Shepard, Roger N and Metzler, Jacqueline},
  journal={Science},
  volume={171},
  number={3972},
  pages={701--703},
  year={1971},
  publisher={American Association for the Advancement of Science}
}

@inproceedings{chen2024spatialvlm,
  title={Spatialvlm: Endowing vision-language models with spatial reasoning capabilities},
  author={Chen, Boyuan and Xu, Zhuo and Kirmani, Sean and Ichter, Brain and Sadigh, Dorsa and Guibas, Leonidas and Xia, Fei},
  booktitle={Proceedings of the IEEE/CVF Conference on Computer Vision and Pattern Recognition},
  pages={14455--14465},
  year={2024}
}

@inproceedings{li2023maskdino,
  title={Mask dino: Towards a unified transformer-based framework for object detection and segmentation},
  author={Li, Feng and Zhang, Hao and Xu, Huaizhe and Liu, Shilong and Zhang, Lei and Ni, Lionel M and Shum, Heung-Yeung},
  booktitle={Proceedings of the IEEE/CVF Conference on Computer Vision and Pattern Recognition},
  pages={3041--3050},
  year={2023}
}

@inproceedings{li2023flip,
  title={Scaling language-image pre-training via masking},
  author={Li, Yanghao and Fan, Haoqi and Hu, Ronghang and Feichtenhofer, Christoph and He, Kaiming},
  booktitle={Proceedings of the IEEE/CVF Conference on Computer Vision and Pattern Recognition},
  pages={23390--23400},
  year={2023}
}

@inproceedings{zhan2024oakink2,
  title={OAKINK2: A Dataset of Bimanual Hands-Object Manipulation in Complex Task Completion},
  author={Zhan, Xinyu and Yang, Lixin and Zhao, Yifei and Mao, Kangrui and Xu, Hanlin and Lin, Zenan and Li, Kailin and Lu, Cewu},
  booktitle={Proceedings of the IEEE/CVF Conference on Computer Vision and Pattern Recognition},
  pages={445--456},
  year={2024}
}

@inproceedings{fan2023arctic,
  title={ARCTIC: A dataset for dexterous bimanual hand-object manipulation},
  author={Fan, Zicong and Taheri, Omid and Tzionas, Dimitrios and Kocabas, Muhammed and Kaufmann, Manuel and Black, Michael J and Hilliges, Otmar},
  booktitle={Proceedings of the IEEE/CVF Conference on Computer Vision and Pattern Recognition},
  pages={12943--12954},
  year={2023}
}

@article{ren2016fasterRCNN,
  title={Faster R-CNN: Towards real-time object detection with region proposal networks},
  author={Ren, Shaoqing and He, Kaiming and Girshick, Ross and Sun, Jian},
  journal={IEEE transactions on pattern analysis and machine intelligence},
  volume={39},
  number={6},
  pages={1137--1149},
  year={2016},
  publisher={IEEE}
}

@article{oquab2023dinov2,
  title={Dinov2: Learning robust visual features without supervision},
  author={Oquab, Maxime and Darcet, Timoth{\'e}e and Moutakanni, Th{\'e}o and Vo, Huy and Szafraniec, Marc and Khalidov, Vasil and Fernandez, Pierre and Haziza, Daniel and Massa, Francisco and El-Nouby, Alaaeldin and others},
  journal={arXiv preprint arXiv:2304.07193},
  year={2023}
}

@inproceedings{
zhu2024languagebind,
title={LanguageBind: Extending Video-Language Pretraining to N-modality by Language-based Semantic Alignment},
author={Bin Zhu and Bin Lin and Munan Ning and Yang Yan and Jiaxi Cui and WANG HongFa and Yatian Pang and Wenhao Jiang and Junwu Zhang and Zongwei Li and Cai Wan Zhang and Zhifeng Li and Wei Liu and Li Yuan},
booktitle={The Twelfth International Conference on Learning Representations},
year={2024},
url={https://openreview.net/forum?id=QmZKc7UZCy}
}

@inproceedings{girdhar2023imagebind,
  title={ImageBind: One Embedding Space To Bind Them All},
  author={Girdhar, Rohit and El-Nouby, Alaaeldin and Liu, Zhuang
and Singh, Mannat and Alwala, Kalyan Vasudev and Joulin, Armand and Misra, Ishan},
  booktitle={CVPR},
  year={2023}
}

@article{shao2023iebins,
  title={Iebins: Iterative elastic bins for monocular depth estimation},
  author={Shao, Shuwei and Pei, Zhongcai and Wu, Xingming and Liu, Zhong and Chen, Weihai and Li, Zhengguo},
  journal={Advances in Neural Information Processing Systems},
  volume={36},
  pages={53025--53037},
  year={2023}
}

@article{li2024binsformer,
  title={Binsformer: Revisiting adaptive bins for monocular depth estimation},
  author={Li, Zhenyu and Wang, Xuyang and Liu, Xianming and Jiang, Junjun},
  journal={IEEE Transactions on Image Processing},
  year={2024},
  publisher={IEEE}
}

@inproceedings{bhat2022localbins,
  title={Localbins: Improving depth estimation by learning local distributions},
  author={Bhat, Shariq Farooq and Alhashim, Ibraheem and Wonka, Peter},
  booktitle={European Conference on Computer Vision},
  pages={480--496},
  year={2022},
  organization={Springer}
}

@inproceedings{zhang2024condense,
  title={ConDense: Consistent 2D/3D Pre-training for Dense and Sparse Features from Multi-View Images},
  author={Zhang, Xiaoshuai and Wang, Zhicheng and Zhou, Howard and Ghosh, Soham and Gnanapragasam, Danushen and Jampani, Varun and Su, Hao and Guibas, Leonidas},
  booktitle={European Conference on Computer Vision},
  pages={19--38},
  year={2024},
  organization={Springer}
}

@inproceedings{
iclr2025depthpro,
title={Depth Pro: Sharp Monocular Metric Depth in Less Than a Second},
author={Alexey Bochkovskiy and Ama{\"e}l Delaunoy and Hugo Germain and Marcel Santos and Yichao Zhou and Stephan Richter and Vladlen Koltun},
booktitle={The Thirteenth International Conference on Learning Representations},
year={2025},
url={https://openreview.net/forum?id=aueXfY0Clv}
}

@inproceedings{
xu2025do,
title={Do Egocentric Video-Language Models Truly Understand Hand-Object Interactions?},
author={Boshen Xu and Ziheng Wang and Yang Du and Zhinan Song and Sipeng Zheng and Qin Jin},
booktitle={The Thirteenth International Conference on Learning Representations},
year={2025},
url={https://openreview.net/forum?id=M8gXSFGkn2}
}

@inproceedings{piccinelli2024unidepth,
  title={UniDepth: Universal monocular metric depth estimation},
  author={Piccinelli, Luigi and Yang, Yung-Hsu and Sakaridis, Christos and Segu, Mattia and Li, Siyuan and Van Gool, Luc and Yu, Fisher},
  booktitle={Proceedings of the IEEE/CVF Conference on Computer Vision and Pattern Recognition},
  pages={10106--10116},
  year={2024}
}

@inproceedings{carion2020detr,
  title={End-to-end object detection with transformers},
  author={Carion, Nicolas and Massa, Francisco and Synnaeve, Gabriel and Usunier, Nicolas and Kirillov, Alexander and Zagoruyko, Sergey},
  booktitle={European conference on computer vision},
  pages={213--229},
  year={2020},
  organization={Springer}
}

@inproceedings{
sam2,
title={{SAM} 2: Segment Anything in Images and Videos},
author={Nikhila Ravi and Valentin Gabeur and Yuan-Ting Hu and Ronghang Hu and Chaitanya Ryali and Tengyu Ma and Haitham Khedr and Roman R{\"a}dle and Chloe Rolland and Laura Gustafson and Eric Mintun and Junting Pan and Kalyan Vasudev Alwala and Nicolas Carion and Chao-Yuan Wu and Ross Girshick and Piotr Dollar and Christoph Feichtenhofer},
booktitle={The Thirteenth International Conference on Learning Representations},
year={2025},
url={https://openreview.net/forum?id=Ha6RTeWMd0}
}

@article{liu2023llava,
  title={Visual instruction tuning},
  author={Liu, Haotian and Li, Chunyuan and Wu, Qingyang and Lee, Yong Jae},
  journal={Advances in neural information processing systems},
  volume={36},
  pages={34892--34916},
  year={2023}
}

@inproceedings{
feng2023layoutgpt,
title={Layout{GPT}: Compositional Visual Planning and Generation with Large Language Models},
author={Weixi Feng and Wanrong Zhu and Tsu-Jui Fu and Varun Jampani and Arjun Reddy Akula and Xuehai He and S Basu and Xin Eric Wang and William Yang Wang},
booktitle={Thirty-seventh Conference on Neural Information Processing Systems},
year={2023},
url={https://openreview.net/forum?id=Xu8aG5Q8M3}
}

@article{zhang2016rgbd,
  title={RGB-D-based action recognition datasets: A survey},
  author={Zhang, Jing and Li, Wanqing and Ogunbona, Philip O and Wang, Pichao and Tang, Chang},
  journal={Pattern Recognition},
  volume={60},
  pages={86--105},
  year={2016},
  publisher={Elsevier}
}

@article{liu2019ntu,
  title={Ntu rgb+ d 120: A large-scale benchmark for 3d human activity understanding},
  author={Liu, Jun and Shahroudy, Amir and Perez, Mauricio and Wang, Gang and Duan, Ling-Yu and Kot, Alex C},
  journal={IEEE transactions on pattern analysis and machine intelligence},
  volume={42},
  number={10},
  pages={2684--2701},
  year={2019},
  publisher={IEEE}
}

@article{zhang2020vslnet,
  title={Span-based localizing network for natural language video localization},
  author={Zhang, Hao and Sun, Aixin and Jing, Wei and Zhou, Joey Tianyi},
  journal={arXiv preprint arXiv:2004.13931},
  year={2020}
}

@inproceedings{zhao2021vsgn,
    title={Video Self-Stitching Graph Network for Temporal Action Localization},
    author={Zhao, Chen and Thabet, Ali K and Ghanem, Bernard},
    booktitle={Proceedings of the IEEE/CVF International Conference on Computer Vision},
    pages={13658--13667},
    year={2021}
  }

@inproceedings{liu2022convnext,
  title={A convnet for the 2020s},
  author={Liu, Zhuang and Mao, Hanzi and Wu, Chao-Yuan and Feichtenhofer, Christoph and Darrell, Trevor and Xie, Saining},
  booktitle={Proceedings of the IEEE/CVF conference on computer vision and pattern recognition},
  pages={11976--11986},
  year={2022}
}

@inproceedings{devlin2019bert,
  title={Bert: Pre-training of deep bidirectional transformers for language understanding},
  author={Devlin, Jacob and Chang, Ming-Wei and Lee, Kenton and Toutanova, Kristina},
  booktitle={Proceedings of the 2019 conference of the North American chapter of the association for computational linguistics: human language technologies, volume 1 (long and short papers)},
  pages={4171--4186},
  year={2019}
}

@article{eigen2014depth,
  title={Depth map prediction from a single image using a multi-scale deep network},
  author={Eigen, David and Puhrsch, Christian and Fergus, Rob},
  journal={Advances in neural information processing systems},
  volume={27},
  year={2014}
}

@article{xue2023learning,
  title={Learning fine-grained view-invariant representations from unpaired ego-exo videos via temporal alignment},
  author={Xue, Zihui Sherry and Grauman, Kristen},
  journal={Advances in Neural Information Processing Systems},
  volume={36},
  pages={53688--53710},
  year={2023}
}

@inproceedings{wang2025vggt,
  title={Vggt: Visual geometry grounded transformer},
  author={Wang, Jianyuan and Chen, Minghao and Karaev, Nikita and Vedaldi, Andrea and Rupprecht, Christian and Novotny, David},
  booktitle={Proceedings of the Computer Vision and Pattern Recognition Conference},
  pages={5294--5306},
  year={2025}
}
}
\clearpage

\appendix









\newpage


We provide additional details, extended experimental results, and further discussion in this supplementary material, including: 
\begin{itemize}[nosep,wide,labelindent=0pt,labelwidth=*,align=left]
    \item Implementation details: background of foundation models (\Cref{subsec:background}), dataset details (\Cref{subsec:datasets}), and experimental details (\Cref{subsec:experimental_details}). 
    \item Additional quantitative results: analyses and evaluations (\Cref{sec:quantitative_results}). 
    \item Additional qualitative results: Examples and visualizations to complement the main results (\Cref{sec:qualitative_results}). 
    \item Discussion on related works: Insights and comparisons with prior research (\Cref{sec:discussion}).
\end{itemize}

\section{Implementation Details}
\subsection{Background of Foundation Models}
\label{subsec:background}
\noindent\textbf{HOI Detector~\cite{hoidetector}.}
HOID is a robust system for detecting human hands and their interacting objects in images.
It is built upon Faster-RCNN~\cite{ren2016fasterRCNN}, pretrained on 100K image dataset with hand-object interaction annotations, including hand bounding box, interacting object bounding box, hand side (left or right), hand contact state (e.g., no contact, self-contact, other person contact, contact with portable object, or contact with a non-portable object).
To enhance its capabilities, HOID is further trained with an additional 42K egocentric data samples, enabling improved understanding of HOI from egocentric view.
We leverage HOID to generate spatial HOI boxes, hand side, hand contact state for each video, from 12 uniformly sampled frames per video. 
However, HOID often produces temporally inconsistent box predictions across adjacent frames. 
To address this, we apply a robust image and video segmentation model to refine the detection results, ensuring greater consistency and accuracy.

\noindent\textbf{Segment Anything 2~\cite{sam2}.}
SAM2 is a versatile segmentation model capable of segmenting objects in both images and videos according to a given prompt, such as a point, box or mask, with remarkable efficiency.
It is trained on a large-scale SA-V dataset, comprising 50.9K videos and 35.5M high-quality masks. SAM2 employs a hierarchical image encoder and a memory mechanism to handle streaming frame input.
In our approach, SAM2 is utilized to generate the spatial-temporal consistent HOI masks by leveraging prompts derived from HOID outputs.


\noindent\textbf{Depth Anything 2~\cite{depthanythingv2}.}
DAv2 excels in monocular depth estimation, offering fine-grained details, strong generalization and efficient inference.
It is built upon the pretrained visual foundation model DINOv2~\cite{oquab2023dinov2} and a depth decoder DPT~\cite{dpt}.
Pretrained on 595K synthetic images and 62M pseudo-labeled real images, DAv2 exhibits strong out-of-domain generalization.
In our work, we use DAv2 to generate depth maps for eight frames sampled from egocentric videos, serving as supervision signals. 
Following the recommendations of DAv2, we employ the DAv2-Large variant, which produces more spatial-temporal consistent depth maps. 

\noindent\textbf{DeepSeek-LLM~\cite{deepseekllm}.}
We employ the LLM to interpret HOI information and enrich textual descriptions with shape and movement details.
DeepSeek-LLM demonstrates exceptional ability to follow instructions and comprehend HOI mask prompts.
Specifically, we use the DeepSeek API to access their 200B-parameter LLM to facilitate these tasks.


\subsection{Dataset Details}
\label{subsec:datasets}
\textbf{Ego4D~\cite{grauman2022ego4d}.}
Ego4D contains 3,670 hours of egocentric videos with dense narrations, covering diverse scenarios and activities from worldwide. 
Each narration is timestamped and paired with a free-form sentence.
Following the approach in ~\cite{lavila}, we construct 4M video-text clip pairs for pre-training, with an average clip length of 1 second (±0.9). 
In our text enrichment process, we only keep those hand-object interaction clips performed by the camera wearer, where the text begins with `\#C` (denoting the wearer) and then follows HOI-related verbs and nouns.
This strategy excludes clips that record other people's activities, such as multi-person interactions~\citep{Ryan2023EgocentricAA} where the text begins with `\#O`, and the videos like `\#C C walks away`.
For the natural language query task, it comprises 1,659 untrimmed videos, each averaging 500 seconds in duration. 
On average, each video contains 12 clip-query pairs. Following the official split from~\cite{grauman2022ego4d}, we use 11,291 queries for training and 3,874 for validation.

\noindent\textbf{Epic-Kitchens~\cite{epic-kitchens-100}.}
Epic-Kitchens-100 (EK-100) consists of 100 hours of egocentric cooking videos divided into training (67,217 clips), validation  (9,668 clips), and testing (13,092 clips) splits. 
Each clip includes start and end timestamps, a short textual narration, and a verb and noun class that correspond to the narration.
There are 3805 action classes, 97 verb classes, and 300 noun classes.
We evaluate our pre-trained model on the validation split. 

\noindent\textbf{EGTEA~\cite{egtea}.}
EGTEA comprises 28 hours of egocentric cooking videos, annotated with 10,321 instances of fine-grained actions across 106 classes. The average action duration is 3.2 seconds.
For our experiments, we use only the visual frames as input. 
We follow prior works~\cite{kazakos2021little,lavila} and report top-1 accuracy and mean class accuracy on all three test splits, including 2,022 testing instances for each split.


\noindent\textbf{H2O~\cite{Kwon2021h2o}.}
H2O is a dataset capturing egocentric hand-object interactions in a laboratory, including 36 action classes. 
The egocentric data is captured from an Azure Kinect camera mounted egocentrically for recordings. 
Since our primary target is to evaluate the transfer learning capability of visual representation, the train/val splits have 7862/11638 frames. 

\subsection{Experimental Details}
\label{subsec:experimental_details}
\noindent\textbf{Zero-Shot Video-Text Retrieval in EK100MIR and EgoMCQ.} 
We perform video-text matching with 16 frames as input for EK100MIR and 4 frames for EgoMCQ, following~\cite{zhao2023avion}.

\noindent\textbf{Zero-Shot Action Recognition in EGTEA. }
We follow the evaluation protocol proposed by~\cite{egtea} to compute the mean performance across all evaluation splits.
This involves performing video-text retrieval between video clips and the action text labels, which are prompted by prepending "\#C C ...".
During inference, we apply three spatial crops of size $224\times 224$ from each $256\times 256$ frame of 10 video clip, averaging predictions across these crops to produce the final results.

\noindent\textbf{Depth Estimation in H2O. }
The frozen visual encoder produces feature maps of dimension 768, which are passed to a linear decoder to estimate depths with a resolution of $720\times 720$.
The model is trained for 10 epochs with a batch size of 64 and a learning rate of 0.0005, where the first 1.5 epochs serve as a warm-up phase. 
Our evaluation code is built upon Probing3D~\cite{cvpr-2024probing3D}.




\noindent\textbf{Robot Manipulation in Franka Kitchen.}
In Franka Kitchen environment, all baselines apply imitation learning for visuomotor control. 
A policy network is trained for each task using observations from the environment and video representation from EgoDTM.
To adapt EgoDTM and LaViLa, we repeat the image observation 4 times as video input.
For each task, the experiment is conducted from two different camera viewpoints for two random seeds using 50 randomly sampled trajectories. The final result is the average of the success rate.
Our codebase is built upon MPI~\cite{zeng2024mpi}.

\noindent\textbf{Natural Language Queries on Ego4D.}
The task typically operates on a 6-minute video.
Using the video compression technique from~\cite{caba2015activitynet}, we compress the original video frames by 6 times to save storage.
Then we extract the features with the fps of 1.87 and sampling frame number 4.
We takes 256 dimension global video features and 768 dimension BERT features as input.
Our codebase is built upon EgoVLP~\cite{lin2022egovlp}.




\section{Additional Quantitative Results}
\label{sec:quantitative_results}

\begin{wraptable}{r}{0.5\linewidth}
\centering
\caption{Multi-modal action recognition on EGTEA using our predicted depths. We apply a simple MLP to encode depth maps and perform multi-modal feature fusion via late fusion.}

\scalebox{0.75}{
\begin{tabular}{c|ccc}
\toprule
Modality & mean-acc & top1-acc & top5-acc  \\
\midrule
RGB & 61.6 & 68.2 & 87.4   \\
RGB+Depth & 62.5 & 69.5 & 89.2 \\
\bottomrule
\end{tabular}
}
\label{tab: analysis_4_RGBD_action_recognition}
\end{wraptable}
\noindent\textbf{Predicted depth contains valuable information for action recognition.}
Since ground-truth data is unavailable for directly evaluating the depth decoder, we demonstrate the utility of our predicted depth maps for multimodal action recognition, as shown in ~\Cref{tab: analysis_4_RGBD_action_recognition}. 
We simply encode the depth map using an MLP at a 56p resolution. 
The action recognition accuracy improved by +1.4\%, confirming that our predicted depths contain meaningful information for unseen egocentric data.

\begin{figure}[t]
  \centering    \includegraphics[width=0.8\linewidth]{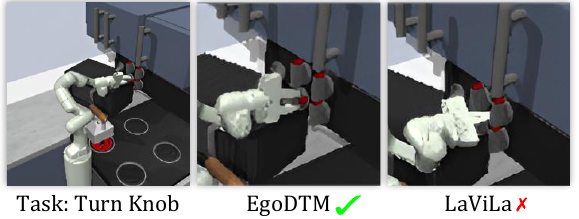}
  \caption{Qualitative results of robot manipulations. }
  \label{fig:case_robot_manipulation}
\end{figure}

\section{Additional Qualitative Results}
\label{sec:qualitative_results}

\begin{figure*}[t]
  \centering
    \includegraphics[width=0.95\linewidth]{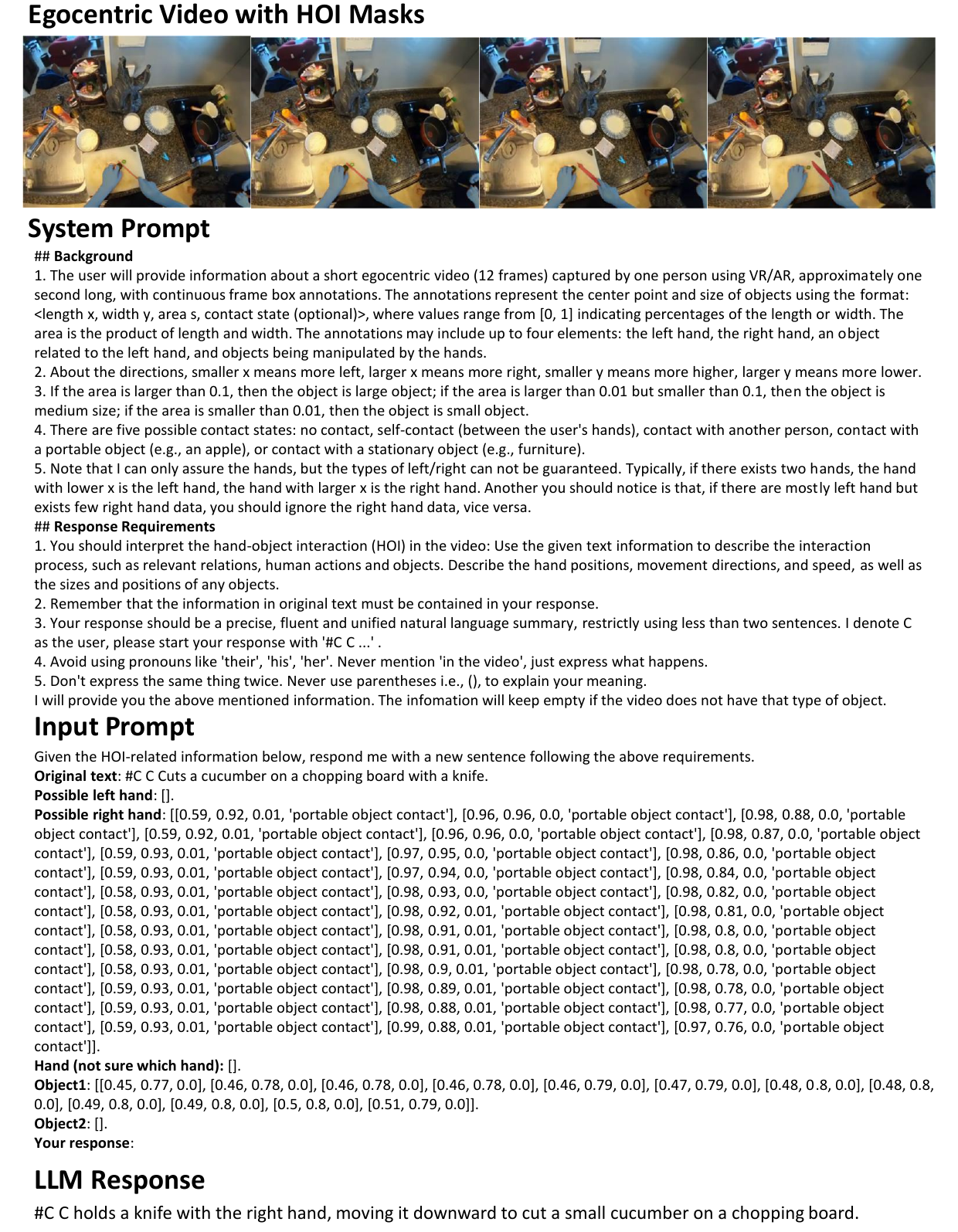}
  \caption{LLM prompt strategy for generating enriched text from HOI masks and the original text.}
  \label{fig:LLM_prompts}
\end{figure*}

\noindent\textbf{LLM Prompts.}
We use the LLM prompts in~\Cref{fig:LLM_prompts} to enrich the texts with HOI shape and movement information.
Specifically, the shape information is provided by the HOI mask areas, where the large, medium, small object occupies [0.1,1], [0.01, 0.1] and [0, 0.01] areas, respectively.

\begin{figure*}[t]
  \centering
    \includegraphics[width=0.95\linewidth]{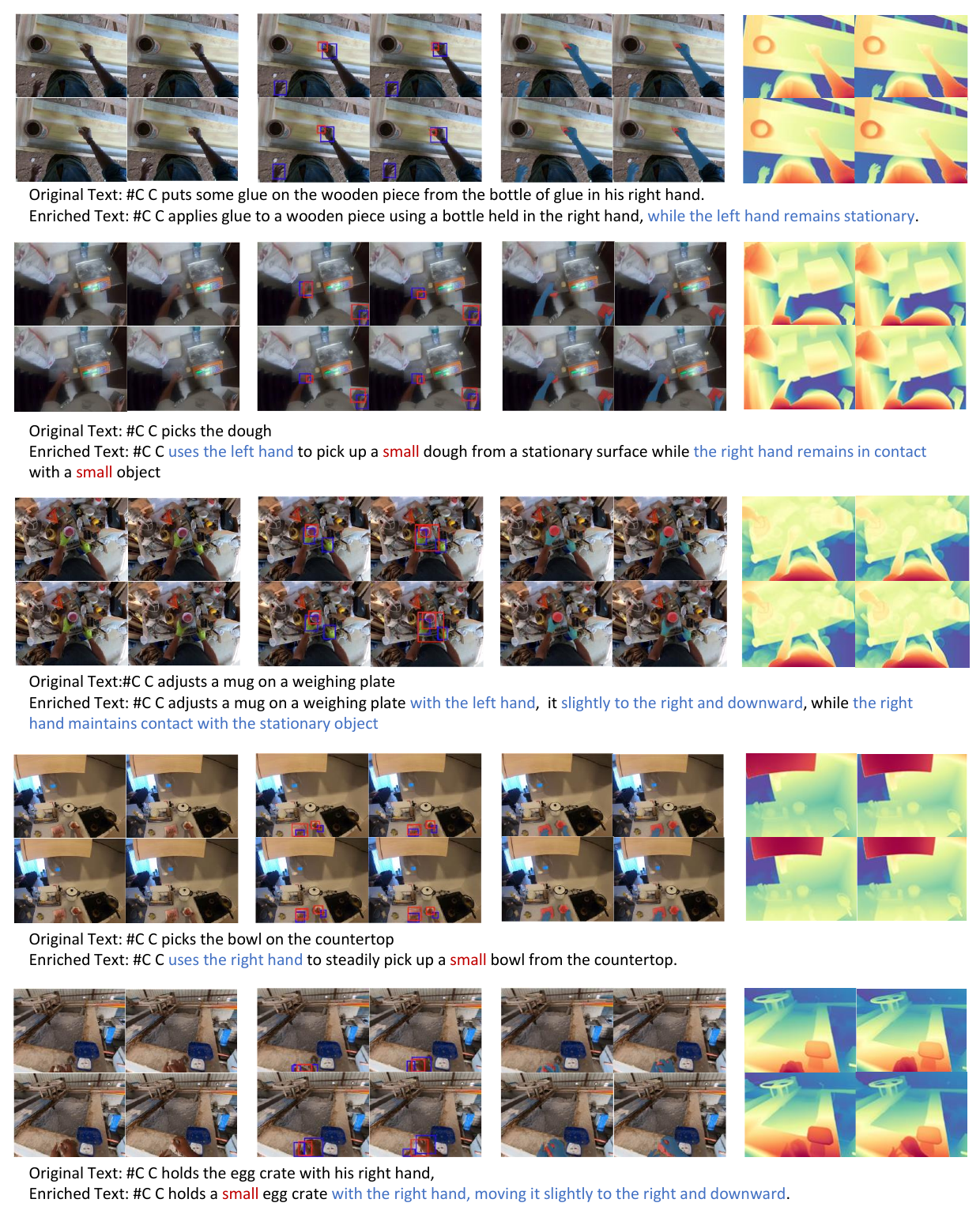}
  \caption{Illustration of data generated by our data generation pipelines, including intermediate HOI boxes, masks, and the enriched texts and depth maps used as supervision signals. The text that includes HOI movements is marked {\color{blue} blue}, while the contents that include HOI spatial information are marked {\color{red} red}. }
  \label{fig:Generated_data}
\end{figure*}

\noindent\textbf{Generated Data.}
Examples of generated HOI boxes, HOI masks, depths, and enriched texts are illustrated in~\Cref{fig:Generated_data}.

\noindent\textbf{Case Study on Robot Manipulation.}
In~\Cref{fig:case_robot_manipulation}, the learned policy based on EgoDTM visual representation enables the robot to approach the switch and turn it, while LaViLa successfully approaches but misses the switch.

\section{Discussions}
\label{sec:discussion}
\noindent\textbf{Comparison with Related Works that Pretrained with Depth. }
ImageBind~\cite{girdhar2023imagebind} and LanguageBind~\cite{zhu2024languagebind} are the most relevant depth-based vision-language pretraining works. 
These methods employ multiple encoders to align various modalities within a unified feature space through contrastive learning.
While both methods utilize depth for pretraining, their application may be less impactful when applied to conventional third-person datasets.
In contrast, depth is essential for egocentric perception, where spatial awareness is critical for understanding human indoor activities.
Furthermore, their pretraining processes treat depth as an input rather than a prediction target, resulting in depth features that lack pixel-level 3D information and video representations that remain unaware of 3D structure.
In our work, we aim to enable video representations to predict depth maps, thereby embedding 3D awareness directly into the representations.

\noindent\textbf{Potential for Real-World Applications.}
While our model demonstrates improvements over text-pretrained models in both video understanding and robotic manipulation tasks, it falls short of state-of-the-art performance of manipulation models.
However, our model predicts more meaningful depth maps in real-world settings than in simulations, offering promising potential for real-world deployment.

\noindent\textbf{}

\end{document}